\documentclass[10pt,journal,compsoc]{article}   %

\usepackage{arxiv}

\usepackage[nolist]{acronym}
\usepackage[version=4]{mhchem}  %
\usepackage{tikz}
\usepackage[decisionutilitycolor]{influence-diagrams}

\newcommand{\NEE}{NEE} 

\newcommand{\GPP}{GPP}
\newcommand{\LUE}{LUE}
\newcommand{\SW}{SW}
\newcommand{\RECO}{R_\text{eco}}
\newcommand{\RECOsyn}{R^{\text{syn}}_\text{eco}}
\newcommand{\Rb}{R_{b}}
\newcommand{\Rbsyn}{R^{\text{syn}}_{b}}
\newcommand{\Q}{Q_{10}}

\newcommand{\VPD}{VPD}

\newcommand{\SWPOTsm}{SW_{\text{POT}}^{\text{SM}}}%
\newcommand{\SWPOTsmdiff}{SW_{\text{POT}}^{\text{SM,diff}}}%

\DeclareMathOperator{\mol}{mol}

\DeclareMathOperator{\E}{\mathbb{E}}

\usepackage{array}   %
\newcolumntype{L}{>{$}l<{$}} %
\newcolumntype{R}{>{$}r<{$}} %

\usepackage{booktabs}
\usepackage{amsfonts}
\usepackage{amssymb}
\usepackage{amsbsy} %
\usepackage{amsmath}
\usepackage{cuted}
\usepackage{flushend}
\usepackage{caption}
\usepackage{subcaption}
\usepackage[T1]{fontenc}
\usepackage[latin1]{inputenc}

\usepackage{hyperref}
\hypersetup{
    unicode=false,          %
    pdftoolbar=true,        %
    pdfmenubar=true,        %
    pdffitwindow=false,     %
    pdfstartview={FitH},    %
    pdftitle={IEEE submission},    %
    pdfauthor={Prof.Fassman},     %
    pdfsubject={RBIG-Information},   %
    pdfcreator={Prof.Fassman},   %
    pdfproducer={Prof.Fassman}, %
    pdfkeywords={keyword1} {key2} {key3}, %
    pdfnewwindow=true,      %
    colorlinks=true,       %
    linkcolor=blue,          %
    citecolor=blue,        %
    filecolor=blue,      %
    urlcolor=blue           %
}

\usepackage{multirow}

\usepackage{amsmath,epsfig}
\usepackage{array}
\usepackage{soul}
\usepackage[normalem]{ulem} %
\usepackage{times}
\usepackage[all]{xy}
\usepackage{flushend}
\usepackage{url}
\usepackage{rotating}
\usepackage{epstopdf}

\usepackage{float}
\usepackage{cite}
\usepackage{balance}
\usepackage{dsfont}

\usepackage{comment}
\usepackage[capitalise]{cleveref}
\usepackage{amssymb}
\usepackage{amsbsy}
\usepackage{amsmath}
\usepackage[most]{tcolorbox}

\usepackage{xcolor}

\begin{acronym}
\acro{DML}[DML]{double machine learning}
\acro{EC}[EC]{eddy covariance}
\acro{GPP}[GPP]{gross primary production}
\acro{LUE}[LUE]{light-use efficiency}
\acro{NEE}[NEE]{net ecosystem exchange}
\acro{NN}[NN]{neural network}
\acro{DNN}[NN]{neural network}
\acro{OOD}[OOD]{out-of-distribution}
\acro{RECO}[RECO]{ecosystem respiration}
\acro{SW}[SW]{incoming shortwaves}
\acro{SWC}[SWC]{soil water content}
\acro{doy}[doy]{day of the year}
\acro{VPD}[VPD]{vapor pressure deficit}
\acro{NT}[NT]{nighttime}
\acro{DT}[DT]{daytime}
\acro{SIF}[SIF]{solar-induced chlorophyll fluorescence}
\acro{RMSE}[RMSE]{root-mean-square error}
\acro{PINN}[PINN]{physics-informed neural network}
\acro{GDHM}[GD-based HM]{gradient-descent-based hybrid modeling}
\acro{DMLHM}[DML-based HM]{DML-based hybrid modeling}
\acro{HM}[HM]{hybrid modeling}
\acro{RF}[RF]{random forest}
\acro{ML}[ML]{machine learning}
\acro{DL}[DL]{deep learning}
\end{acronym}

\usepackage{cite}

\begin{document}
\title{Causal hybrid modeling with double machine learning}

\author{
  Kai-Hendrik Cohrs\\
  Image Processing Laboratory \\
  Universitat de Val{\`e}ncia\\
  Val{\`e}ncia, Spain\\
  \texttt{kai.cohrs@uv.es} \\
  \And
  Gherardo Varando\\
  Image Processing Laboratory \\
  Universitat de Val{\`e}ncia\\
  Val{\`e}ncia, Spain\\
  \texttt{gherardo.varando@uv.es} \\
  \And
  Gustau Camps-Valls\\
  Image Processing Laboratory \\
  Universitat de Val{\`e}ncia\\
  Val{\`e}ncia, Spain\\
  \texttt{gcamps@uv.es} \\
  \And
  Nuno Carvalhais\\
  Max Planck Institute for Biogeochemistry\\
  Jena, Germany\\
  ELLIS Unit Jena\\
  \texttt{ncarvalhais@bgc-jena.mpg.de}\\
  \And
  Markus Reichstein\\
  Max Planck Institute for Biogeochemistry\\
  Jena, Germany\\
  ELLIS Unit Jena\\
  \texttt{Markus.Reichstein@bgc-jena.mpg.de} \\
}
\maketitle

\begin{abstract}
Hybrid modeling integrates machine learning with scientific knowledge 
to enhance interpretability, generalization, and adherence to natural laws. Nevertheless, equifinality and regularization biases pose challenges in hybrid modeling to achieve these purposes. This paper introduces a novel approach to estimating hybrid models via a causal inference framework, specifically employing Double Machine Learning (DML) to estimate causal effects. We showcase
its use for the Earth sciences on two problems related to carbon dioxide fluxes. In the $\Q$ model, we demonstrate that DML-based hybrid modeling is superior in estimating causal parameters over end-to-end deep neural network (DNN) approaches, proving efficiency, robustness to bias from regularization methods, and circumventing equifinality. Our approach, applied to carbon flux partitioning, exhibits flexibility in accommodating heterogeneous causal effects. The study emphasizes the necessity of explicitly defining causal graphs and relationships, advocating for this as a general best practice. We encourage the continued exploration of causality in hybrid models for more interpretable and trustworthy results in knowledge-guided machine learning.
\end{abstract}

\vspace{2pc}

\newpage
\section{Introduction}
\label{sec:introduction_PKL}

\noindent 
\Ac{ML}, specifically \ac{DL}, has proven to be effective in identifying and modeling complex patterns from data sets. This led to unprecedented progress in fields such as computer vision\cite{kirillov2023segment}, natural language processing~\cite{brown2020language}, and speech recognition~\cite{zhang2022pushing}. These data-driven models also increasingly complement or even substitute mechanistic methods in science~\cite{Halevy09, Lipton18}.

In the Earth sciences, for instance, the common way to understand and model the Earth's properties, structure, and processes is using knowledge of first principles, realized in mechanistic models based on functional equations~\cite{kump2013earth}. These models allow principled predictions of how the system under study would behave under different conditions~\cite{O'Neill2016}. Nevertheless, they are not always sufficient to capture the complex and usually not completely known relationships in the real world. 

Computational constraints and missing understanding have led to simplified or even missing representation of important processes in the current generation of climate models~\cite{Eyring2016}. Structural limitations often necessitate parameterizations to approximate complex processes. Significant uncertainties include the representation of cloud feedbacks~\cite{Myers2021}, resolving ocean components at varying resolutions~\cite{Hewitt2020}, surface energy partitioning~\cite{YUAN2022108920}, representing key processes like vegetation response to \ce{CO2}~\cite{Arora2020}, and difficulties in representing functional structures across different biome types~\cite{Zhu2014}. Addressing these challenges is essential for enhancing the accuracy and reliability of Earth system models in projecting future climate change and weather extremes.

Integration of machine learning (ML) with abundant Earth data presents a promising avenue to overcome the limitations of current Earth system models~\cite{reichstein19nat,Camps-Valls2021}. Support vector machines~\cite{CampsValls09wiley}, \acp{RF}\cite{Tramontana16bg}, or \acp{DNN}\cite{CampsValls21wiley} are highly flexible, make little prior assumptions on the functional form and can integrate the large datasets abundant in Earth and climate sciences.

The flexibility of \ac{ML} models comes with some known downsides: (i) Many popular machine learning models are black boxes, meaning that we do not understand the internal reasoning behind the model's predictions~\cite{Rudin2019Why}. 
(ii) Often, \ac{ML} models are not robust and fail to generalize out of the domain of the data used for training~\cite{QuioneroCandela2009, 6278037}. 
(iii) They violate physical properties and laws of nature, such as conservation laws, symmetries, or equi- and invariances~\cite{reichstein19nat,marcus2018deep}. 
These are crucial matters in Earth and climate sciences, where a prime goal is to make realistic predictions on the Earth's system under a changing climate~\cite{IPCC_WG1}.

All these issues are gaining attention in \ac{ML} and Earth system science literature. Research in generalization and extrapolation aims at ensuring robustness outside of the training domain~\cite{Neyshabur2017, Wang2022, shen2023engression}.
Explainable artificial intelligence (XAI) tackles questions on the explainability of black box models~\cite{Roscher2020, Linardatos2021, ras2022}, which find growing usage in remote sensing problems~\cite{Mamalakis2022, hohl2024}. At the same time, the general goal of explaining black boxes is being challenged by advocates for glass box models, i.e., inherently interpretable models~\cite{Rudin2019, Rudin2022}, and there is an ongoing debate on the evaluation and rigorousness of XAI methods~\cite{Sixt2020, freiesleben2023dear}.

A flourishing area of research is science-aware or knowledge-guided machine learning (KGML), which combines the knowledge-driven and data-driven worlds to overcome inconsistencies~\cite{karpatne2022knowledge}. These methods increasingly find their way into various domains within Earth sciences~\cite{CampsValls18sciasi, Tramontana2020, khandelwal2020, Cortes21fkl, Liu2023, Zhu2022}. One example is \acp{PINN}~\cite{raissi2019}, where an additional term is added to the loss for training that punishes deviations from physical laws encoded with ODEs or PDEs. Alternatively, \ac{ML} models can be trained on a combination of data and simulations from physical models to improve consistency in the sparse observation regime~\cite{CampsValls18sciasi}. 

Finally, hybrid modeling replaces some components of mechanistic models with machine learning~\cite{zhao2019physics, Reichstein2022, Koppa2022}. This constraint makes the models more interpretable and serves as a regularizer for better generalization to unseen data. If we use deep learning models as the machine learning component, the only requirement for fitting these hybrid models is that the parametric components are differentiable~\cite{Shen2023}. Then, gradient-based optimization allows joint optimization of the \ac{NN} parameters and physical parameters of the mechanistic model and leads to seamless data integration.
In the following, we will refer to this as \textit{\ac{GDHM}}. It serves as a baseline for our proposed method.

There are persisting challenges in hybrid modeling. Firstly, these models are prone to \emph{equifinality}, which denotes the existence of multiple models and sets of parameters that describe the data similarly well. Already in the standard mechanistic modeling, this is a well-known difficulty when not only model performance but also retrieving meaningful parameters is the goal. In this setting, robust inference already poses a challenge~\cite{Oberpriller2021}, which becomes even more difficult and prohibitively expensive in deep learning~\cite{Abdar2021, Izmailov2021WhatAB}. Ultimately, equifinality can jeopardize the interpretability of the results. Second, regularization techniques in machine learning can introduce bias on the physical parameters~\cite{Reichstein2022}. Finally, given the flexibility of non-parametric models such as \acp{DNN}, it is tempting to use different sets of variables for the model and choose the ones that lead to the best overall performance. For a pure prediction task, that is a sensible procedure~\cite{Kuhn2013}. For hybrid modeling, though, apart from equifinality, this can lead to bias or different interpretations of the parameter of interest in the causal sense. We might be \textit{right for the wrong reasons} and imperil the desired interpretability of the hybrid model (see Box \ref{box: equifinality} for an illustrative example).

In many instances, physical equations encode actual cause-effect relationships. It is essential to capture the causal relationships between the variables to obtain interpretable and more accurate models. Respecting the causal direction of time has shown to be effective in training \acp{PINN} for chaotic systems where previous approaches failed~\cite{wang2022respecting}. Furthermore, coupling causal discovery to identify the causal drivers in climate models before applying deep learning algorithms improved performance and interpretability~\cite{iglesiassuarez2023causallyinformed, Runge19}. Causally constrained recurrent \acp{NN} more accurately reflect underlying processes and were shown to enhance our understanding of methane in wetlands~\cite{YUAN2022109115}. Ultimately, causality aims at \textit{being right for the right reasons}.

Therefore, we believe it is time for a {\em causal hybrid modeling} framework, where we introduce an explicit physical prior by assuming a causal graph and framing the problem as a causal effect estimation problem within the hybrid modeling framework. We will show how this approach leads to well-defined problems, thus mitigating equifinality and being robust to biases of training and regularization. As a first step, we propose a method based on \ac{DML}~\cite{Chernozhukov2018}. \ac{DML} is a causal effect estimation technique developed in econometrics, where it is common to investigate the effect of some proposed treatment on an outcome variable~\cite{Knaus_2020, Davis2017}. It has recently been used for effect estimation in the environmental sciences~\cite{SUN2023}.
We suggest that this causal effect estimation technique can be applied to a class of hybrid models where the effect of some input driver on the output is encoded. We coin this method \emph{\ac{DMLHM}}.

Apart from the causal perspective, \ac{DML} has favorable properties over naive fitting approaches. Regularization of the estimators for the non-parametric part of the equation can introduce substantial bias in estimating the parametric part of the equation. Using \ac{DML}, even for erroneous estimators, we can still obtain consistent estimators of the causal effect coefficient. This is particularly useful if the confounding effects are high-dimensional or are described by a complicated function that is hard to learn. Furthermore, it enables us to make inferences, as the estimators are shown to be approximately normally distributed, which yields confidence intervals~\cite{Chernozhukov2018}.

Within the proposed framework based on \ac{DML}, we can solve problems that can be transformed into a regression problem of the form
\begin{center}
    \fbox{
    \begin{minipage} 
    {\dimexpr 10cm}
    \begin{align}\label{eq: problem}
        Y = \theta(X)\cdot f(T) + g(X, W),
    \end{align}
    \vspace{-0.25cm}
    \end{minipage}
    }
\end{center}
where $T$ is a one-dimensional input variable and $X$ and $W$ are further sets of predictors. We assume that $f$ is a known transformation of $T$, and our hybrid modeling goal is to estimate the non-parametric functions $\theta$ and $g$. We will see relevant examples of problems that fall into this class. This includes, in particular, the problems where $\theta$ describes the effect of $T$ on $Y$. This effect can be constant or depend on some other predictors $X$.

We demonstrate the advantages of \ac{DMLHM} in two examples around carbon fluxes:
\begin{enumerate}
    \item The temperature sensitivity $\Q$ model for ecosystem respiration~\cite{arrhenius1889reaktionsgeschwindigkeit, van1899lectures, lloyd1994temperature} and,
    \item the light-use efficiency model for carbon flux partitioning~\cite{PEI2022}.
\end{enumerate}
These two models are particularly relevant as they allow statements on the productivity and respiration of plants under changing conditions.

Our contributions are as follows: In the case of synthetic data for $\Q$, \ac{DML} retrieves the $\Q$ temperature sensitivity parameter more robustly and efficiently than the \ac{GDHM} approach, especially in the low data regime and under regularization. It retrieves $\Q$ values consistent with the literature on measured respiration data. We show how equifinality can yield misleading results and how causal prior knowledge can solve the problem without giving up flexibility. %
In the carbon flux partitioning problem, we show how the method can be extended to the non-linear heterogeneous case, where the hybrid modeling retrieves consistent fluxes and shows competitive performance to the current state-of-the-art neural network.

In essence, we introduce \ac{DMLHM} as a novel approach to fitting hybrid models and show that the obtained estimates are more efficient and robust than the ones from \ac{GDHM}. We describe a path to better pose problems with equifinality, enforcing causal interpretability instead of hoping for it.

\begin{tcolorbox}[colback=black!5!white, colframe=black!50!white, title=Box 1: Equifinality in hybrid modeling, sharp corners, arc=5mm]
\label{box: equifinality}
Modeling the temperature dependence of ecosystem respiration $\RECO$ is a fundamental step in better understanding biosphere evolution and responses under global warming scenarios~\cite{kirschbaum2000will, smith2013plant, huntingford2017implications}. The functional relationship between temperature and respiration has been classically represented via the $\Q$ respiration model:
\begin{align}\label{eq:Q10e}
    R_{eco}(X,T_A)=R_b(X, T_A)\cdot \Q^{(T_A-T_{A}^{ref})/10},
\end{align}
where $\Q$ is the parameter describing temperature sensitivity, $X$ is a set of meteorological drivers and $R_b$ describes the base respiration. 
Including air temperature $T_A$ as a driver of $R_b$ is an optional choice if we are to believe that there are effects of temperature beyond the exponential dependency through $\Q$. A common hybrid modeling approach amounts to using a \ac{DNN} as an estimator for $R_b$, treating $\Q$ as a trainable parameter, and fitting everything end-to-end with gradient descent, as it has been done in~\cite{Reichstein2022}.

Equifinality in this problem can be shown by reformulating~\eqref{eq:Q10e} for $c>0$:
\begin{align}\label{eq:Q10c}
    R_{eco}(X,T_A)=R_b(X, T_A)c^{(T_A-T_{A}^{ref})/10}\cdot \left(\frac{\Q}{c}\right)^{(T_A-T_{A}^{ref})/10}.
\end{align}
Thus, a flexible enough function estimator (e.g. a \ac{DNN}) could learn $R_b(X, T_A)c^{(T_A-T_{A}^{ref})/10}$ and obtain $\frac{\Q}{c}$ as the temperature sensitivity. 
In this case, we would obtain one of the solutions by chance and thus reach erroneous conclusions about the temperature sensitivity.

In this example, equifinality arises because the problem is mathematically ill-posed. It is less obvious, however, when introducing several non-parametric models in more complicated physical equations. In practice, we will obtain a distribution over the parameters mainly driven by inductive biases of the learning algorithm or the network architecture~\cite{Vardi2023} and which are not guided by any physical knowledge.
Additional explicit information can alleviate this problem. These include the introduction of additional losses or adding prior knowledge~\cite{Zhan2022, ElGhawi_2023}. Similarly, a regularization term can make the problem identifiable. This has been formally proven for solving hybrid ODEs~\cite{Yin_2021}. Regularization, however, is known to introduce bias on parameters of interest in semi-parametric modeling problems\cite{Chernozhukov2018}. 
\end{tcolorbox}

\section{Double machine learning for hybrid modeling -- a causal perspective}
\label{sec:methods_PKL}

\begin{figure}
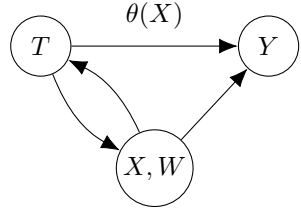

    \centering
    \begin{influence-diagram}
      \node (help) [draw=none] {};      
      \node (T) [left = of help] {$T$};
      \node (Y) [right = of help] {$Y$};
      \node (X) [below = of help] {$X, W$};
      \draw node[midway, above, draw=white, shape=rectangle] {$\theta(X)$};
      \path (X) edge[->, bend right=20] (T);
      \path (T) edge[->, bend right=20] (X);
      \edge {T} {Y};
      \edge {X} {Y};
    \end{influence-diagram}
  \caption{Causal graph of treatment effect estimation of $T$ on $Y$. Sets $X$ and $W$ can enter both as confounders and mediators. Treatment effect $\theta$ can be heterogeneous and dependent on $X$ or constant.}
  \label{fig: causal graph}
\end{figure}

Our setting considers problems that can be expressed as in \eqref{eq: problem}, which can be studied under a causal perspective, see~\cref{fig: causal graph}. The parameter $\theta$ describes the direct effect of some treatment variable $T$ on the outcome variable $Y$. Moreover, we have access to sets of predictors $X$ and $W$ that are confounding or mediating the effect of $T$ on $Y$. Confounders are common causes of $T$ and $Y$, while mediators are variables through which $T$ indirectly affects $Y$. 
The inclusion of mediators has important implications for the interpretation of the results. When we estimate the effect of $T$ on $Y$ with mediators, we only obtain the direct effect by discounting the effects through these mediators. The variables in $X$ can further enter as effect modifiers by modulating the effect $\theta$ of $T$ on $Y$. Technically, we can use all mediators and confounders as effect modifiers when we include them all in $X$, leaving $W$ empty, or treat $\theta$ as a constant effect by instead leaving $X$ empty. At this point, we need to be careful with the choices of control variables $X$ and $W$ as we need to assume that all relevant confounders are observed and included. In particular, this means we need to be careful not to include mediators that have an unobserved common cause with $Y$ or that we introduce a common effect of $T$ and $Y$. Both cases would open a new path and substantially bias the estimation~\cite{Huenermund2021}.

As per the \ac{DML} framework, we must define an auxiliary equation that models the confounding and mediating effects of $X$ and $W$ on $T$. 
Assuming, without loss of generality, centered noise for both equations, we obtain
\begin{align}
Y &= \theta(X)\cdot f(T) + g(X,W)+\epsilon \; &\E[\epsilon|X, W]=0\label{eq:semilin}\\
f(T) &= m(X,W) + \eta & \E[\eta|X, W]=0\\
 & & \E[\eta\cdot \epsilon |X,W]=0.
\end{align}
Sometimes, the original problem formulation must be manipulated to fit our setting. We will see examples of given transformations $f$, though the identity $f(T)=T$ could also be used when the relationship is assumed linear in $T$. The \emph{causal effect} $\theta$ is modeled either as a constant coefficient or as a function of some covariates (heterogeneous effect).

We proceed according to the {\em partialling out} method in the \ac{DML} framework~\cite{Chernozhukov2018}:
\begin{enumerate}
    \item Fit an estimator $\E[Y|X,W]$ of $Y$ on $X$ and $W$,
    \item fit an estimator $\E[f(T)|X,W]$ of $f(T)$ on $X$ and $W$,
    \item compute their residuals as
    $Y_{res} = Y-\E[Y|X,W]$ and $f(T)_{res} = f(T)-\E[f(T)|X,W]$ and
    \item estimate $\hat{\theta} = \arg\!\min_{\theta \in \Theta} \E_n \left[(Y_{res} - \theta(X)\cdot f(T)_{res})^2\right]$.
\end{enumerate}

We call the estimators in (i) and (ii) the first-stage estimators. 
The primary benefit of the \Ac{DML} framework is that it yields fast estimation rates and, under certain assumptions, asymptotic normality of $\theta$. It is robust to errors in the first-stage estimators due to overfitting or regularization bias. This robustness stems from the observation that the moment equations corresponding to the final least squares loss in (iv) fulfill Neyman orthogonality with respect to the first-stage estimators~\cite{Chernozhukov2018}. This approach has been analyzed for a large set of model classes\cite{Chernozhukov2018, Athey2019, Nie2020, Foster2020, nekipelov2021regularized}. For example, any combination of linear regression, decision trees, support vector machines, or \acp{DNN} can be used to model the treatment and/or the outcome models. 
Similarly, any of these or a combination of models could be chosen to estimate the treatment effect. 
To maintain the theoretical guarantees of the \ac{DML} framework, it is important to split the data and perform the first two fitting steps ((i),(ii)) on a different data subset than the last fitting step for the residuals (iv). By doing cross-fitting, data efficiency can be maintained.

\begin{figure}
    \centering
     \usetikzlibrary{arrows.meta}
\usetikzlibrary{backgrounds}
\begin{tikzpicture}[node distance=1cm,font=\scriptsize]
  \tikzstyle{block} = [rectangle, draw, text width=8em, text centered, rounded corners, minimum height=4em, fill=gray!15]
  \tikzstyle{line} = [draw, -{Latex[length=2mm]}]

  \node [block, minimum width = 12em, text width=12em, minimum height=11em, text depth = 9em, opacity = 0.5, text opacity=1] (causal) {{\bf Causal frame}\\[1ex] $
  \begin{array}{l} 
      \begin{influence-diagram}
      \node (help) [draw=none] {};      
      \node (T) [left = of help] {$T$};
      \node (Y) [right = of help] {$Y$};
      \node (X) [below = of help] {$X, W$};
      \draw node[midway, above, draw=white, shape=rectangle] {$\theta(X)$};
      \path (X) edge[->, bend right=20] (T);
      \path (T) edge[->, bend right=20] (X);
      \edge {T} {Y};
      \edge {X} {Y};
    \end{influence-diagram}
  \end{array}$
  \vspace{-1.5em}
  \begin{align*}
  Y = \theta(X) \cdot f(T) + g(X, W)      
  \end{align*} };
  
  \node [block, text width=15em, minimum width = 15em, minimum height=11em, text depth = 9em, opacity = 0.5, text opacity=1] (doubleml) at (5.5, 0) {\bf \bf Double ML\\[1ex] 
  \begin{align*}
  Y_{res} &= Y - \mathbb{E}[Y|X,W] \\ 
  f(T)_{res} &= f(T) - \mathbb{E}[f(T)|X,W]
  \end{align*}
  $\begin{array}{l}
  \hat{\theta} =  \arg\!\min\limits_{\substack{\theta \in \Theta}}\mathbb{E}_n \left[ (Y_{res} - \theta(X) \cdot f(T)_{res})^2 \right]\end{array}$
  };
  
  \node [block, minimum height=11em, text depth=9em, opacity = 0.5, text opacity=1, text width=15em, minimum width = 12em] (estimate) at (11.5,0) {\bf Estimate\\[1ex] 
   \hspace{-2em}
  \vspace{0.2em}
  \begin{itemize}
      \item[(i)] plug-in: 
      \begin{flalign*} \hat{g}(X, W) =& \mathbb{E}[Y|X, W] \\ &- \hat{\theta}(X)\mathbb{E}[f(T)|X,W] \end{flalign*}
      \vspace{-2.2em}
      \item[(ii)] refit:\\
      $\begin{array}{l} \text{build } \hat{g} \text{ on residuals }
      \end{array}$
      \vspace{-2pt}
      \begin{align*}
      Y - \hat{\theta}(X)f(T)
      \end{align*}
  \end{itemize}
 };
  
  \node [block, minimum width=20em, text width=20em, opacity = 0.5, text opacity=1] (combine) at (5.5,-3.5) {\bf Combine\\[1ex] $\hat{Y}(X, W, T) = \hat{\theta}(X)f(T)+\hat{g}(X, W)$};

      \begin{scope}[on background layer]
\draw[-{Triangle[width=48pt,length=18pt]}, line width=30pt, lightgray, shorten >=-20pt, shorten <=-20pt](causal) to (doubleml);

  \draw[-{Triangle[width=48pt,length=18pt]}, line width=30pt, lightgray, shorten >=-20pt, shorten <=-20pt] (doubleml) -- (estimate);
  
\path[-{Triangle[width=48pt,length=18pt]}, line width=30pt, lightgray, shorten >=-15pt, shorten <=-15pt] (estimate.south) edge [ out = 270, in = 0] (combine.east);

\path[-{Triangle[width=48pt,length=18pt]}, line width=30pt, lightgray, shorten >=-15pt, shorten <=-15pt] (causal.south) edge [ out = 270, in = 180] (combine.west);

\path[-{Triangle[width=48pt,length=18pt]}, line width=30pt, lightgray, shorten >=-15pt, shorten <=-15pt] (doubleml.south) edge [ out = 270, in = 90] (combine.north);
\end{scope}
\end{tikzpicture}
   \caption{
Schema of the proposed approach:
   (i) \textbf{Frame} the problem as a
treatment effect estimation
problem and assume causal
graph. (ii) Build estimators of $Y$ and $f(T)$
and deploy \textbf{DML} in the constant or
heterogeneous treatment effect
setting. 
(iii) \textbf{Estimate} $g$ with plug-in estimator or
via a final fitting on the
residuals.
And finally, (iv) \textbf{Combine} $\hat{\theta}$ and $\hat{g}$ into a causally
interpretable hybrid model.}
  \label{fig: schema}
\end{figure}
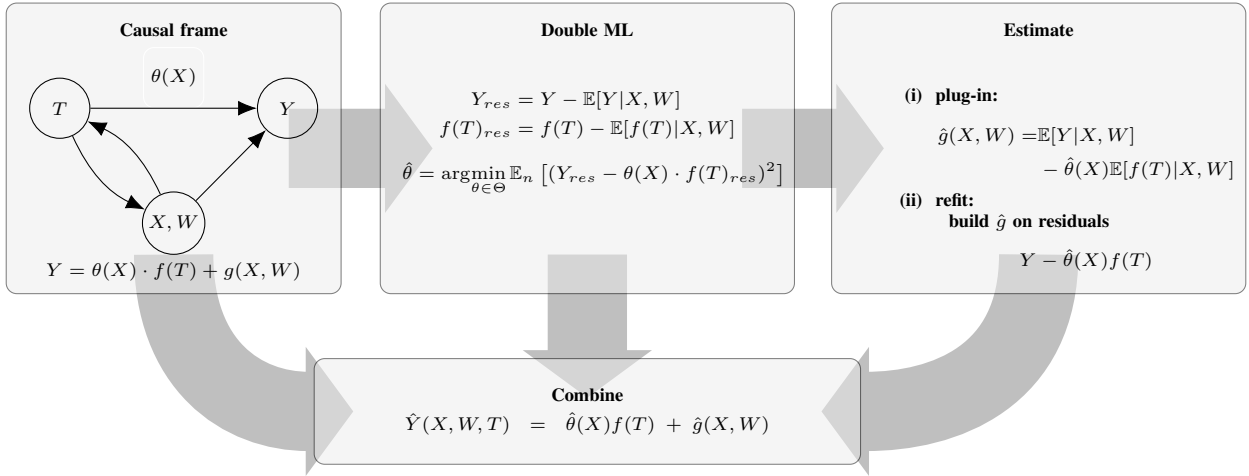

If the only object of the analysis is the interpretable treatment effect $\theta$, the task is completed by the above \ac{DML} procedure. Nevertheless, as is usually the case in hybrid modeling tasks, we are probably also interested in obtaining an estimator of $g$. For this, we have two options:
\begin{enumerate}
    \item Use $\hat{g}(X,W)=\E[Y|X,W] - \hat{\theta}(X)\cdot \E[f(T)|X,W]$ (plug-in) or 
    \item build an estimator on the residuals $Y-\hat{\theta}(X)\cdot f(T)$ (refit).
\end{enumerate}
The plug-in estimator (i) uses all estimators fitted in the previous steps 
and can be obtained at no additional computational cost. A derivation of this estimator is given in \cref{sec: g estimator}.
On the downside, in contrast to $\theta$, there are no theoretical guarantees on how well it describes $g$. 
Option (ii) adds a final supervised learning step, with the advantage being that we are not limited to using the $X$ and $W$ to estimate $\theta$. 
Once $\theta$ has been estimated in a well-posed setting, we can now introduce, for example, $T$ as a driver in the estimation of $g$. 
We can combine all estimators to obtain the fitted hybrid model for \cref{eq: problem} (see \cref{fig: schema} for a summary of the proposed procedure). By separating the problem into a causal inference and a standard supervised learning step, we have maintained its well-posedness. %
Next, we will explain how this technique can be effectively applied in two use cases around carbon fluxes.

\section{Case studies}
\label{sec:data_PKL}
Carbon fluxes are crucial in the global carbon cycle, a key component of the Earth's climate system~\cite{Bonan_2015}. Net ecosystem exchange $NEE$ is the net carbon dioxide flux measured using the \ac{EC} technique~\cite{Burba2013}. The data for our studies is half-hourly data from FLUXNET, a global network of \ac{EC} towers that collect data on carbon dioxide, energy fluxes, sensible heat fluxes, and water vapor exchange between the atmosphere and the terrestrial biosphere~\cite{Fluxnet}. It offers comprehensive measurements of meteorological parameters and constitutes a crucial data source for ecosystem modeling and climate research.

Different biogeochemical processes contribute to the carbon balance of the land~\cite{Falge2003}. In particular and as common, we split $NEE$ as
\begin{align}\label{eq:nee}
    NEE = -GPP + \RECO,
\end{align}
where gross primary production $GPP$ describes the gross carbon uptake by the environment and ecosystem respiration $\RECO$ denotes the carbon release of all organisms.

\subsection{The $\Q$ model}
\label{sec:dataq10}
A common parametrization of $\RECO$ is the $\Q$ respiration model~\cite{arrhenius1889reaktionsgeschwindigkeit, van1899lectures, lloyd1994temperature}:
\begin{align}\label{eq:Q10}
    R_{eco}(X,T_A)=R_b(X)\cdot \Q^{(T_A-T_{A}^{ref})/10}.
\end{align}
This model highlights temperature $T_A$ as a principle driver of respiration, with $\Q$ denoting the temperature sensitivity parameter. Furthermore, $R_b$ describes the base respiration, and $X$ a set of meteorological drivers.
Following the example of~\cite{Reichstein2022}, we use data from the EC tower in Neustift, Austria, available in the FLUXNET2015 dataset~\cite{Pastorello2020}. Based on this site, we extensively probe the \ac{DMLHM} in the controlled setting of synthetic data and showcase its potential on measured data. 
As the goal of this paper is not to provide a comprehensive analysis of global $\Q$ values, we limit ourselves to this site for our first use case.

\paragraph{Data}
Synthetic data is generated from a $\Q$ model with seasonally varying base respiration and measured air temperature $T_A$, and with true constant $\Q$ set to $1.5$ (for details, see \cref{sec: syn Q10 model}).
We provide additional experiments for $\Q$ values of $1.25$ and $1.75$ to showcase the robustness of the results.

Ecosystem respiration is a latent flux not directly observed at flux towers during the day. It can only be measured as nighttime $NEE$, as without photosynthesis, we assume $GPP$ to be zero or under controlled conditions like a sealed chamber~\cite{Falge2003}. We use 2003 to 2007 for training and keep 2008 and 2009 for testing. 
Moreover, we consider only measured observations, which amount
to approximately $10\%$ of the nighttime data for training (4331 data points).

\paragraph{Applying \ac{DMLHM}}
Applying a $\log$-transform to \eqref{eq:Q10} and setting $f(T_A) = (T_A-T_{A}^{ref})/10$ yields
\begin{align}\label{eq:logQ10}
    \log(R_{eco}(X,T_A))=\log(R_b(X)) +  f(T_A)\cdot \log(\Q).
\end{align}
The resulting equation \eqref{eq:logQ10} describes a partially linear regression problem~\cite{Robinson1988} equivalent to~\eqref{eq: problem}. Here, $\log(\Rb(\cdot))$ represents the non-parametric function $g(\cdot)$ as we do not know the functional form of $\Rb$. We aim to estimate the constant linear effect $\theta = \log(\Q)$ of the transformed temperature $f(T_A)$ on the log-transformed ecosystem respiration.
In this work, we employ and compare both \acp{DNN} and \acp{RF} as examples for first-stage estimators. 

After obtaining the estimator $\hat{Q}_{10}$, we fit a \ac{DNN} on 
\begin{align}\label{eq:residuals}
    \frac{R_{eco}(X,T_A)}{\hat{Q}_{10}^{f(T_A)}}=NN(X,T_A).
\end{align}
We compare the {\em causal \ac{DMLHM}} to the {\em standard \ac{GDHM}} as described in \cite{Reichstein2022}. We fit
\begin{align}
     R_{eco}(X,T_A)=NN(X)\cdot \Q^{(T_A-T_{A}^{ref})/10},
\end{align}
with a \ac{DNN} representing the base respiration $\Rb$.
The weights of the \ac{DNN} are optimized together with $\Q$ using the Adam~\cite{kingma2017} optimizer. 

We run the experiments with and without regularization for all involved \acp{DNN} in both hybrid modeling approaches. For this, we use dropout at a rate of $0.2$. This technique randomly drops connections in a \ac{DNN} during training and was found to have a sparsifying effect on the model~\cite{Srivastava2014}. 
We provide additional experiments with weight decay~\cite{Krogh1991}, another common regularization technique in deep learning at a rate of $0.1$. To showcase the effect of equifinality, we also introduce $T_A$ as an additional predictor in $\Rb$. 
We will apply the same training procedure and \ac{DNN} architectures for both hybrid modeling approaches for comparability and to show robustness in the presence of biased estimators. We only drop the final nonlinearity for the first-stage estimators in the \ac{DMLHM}. Details on the \acp{DNN} and their training can be found in \cref{sec: model details}.

\begin{figure}
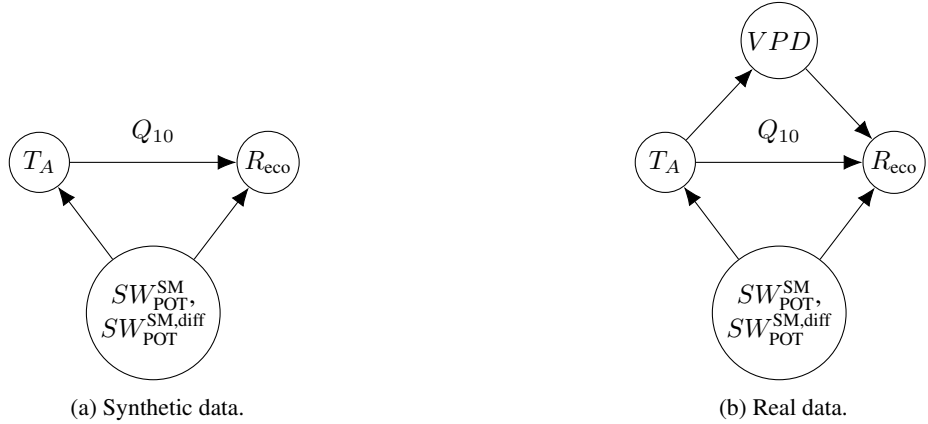

  \begin{subfigure}{0.5\textwidth}
    \centering
    \begin{influence-diagram}
      \node (help) [draw=none] {};      
      \node (T) [left = of help] {$T_A$};
      \node (Y) [right = of help] {$\RECO$};
      \node (X) [below = of help] {$\SWPOTsm$,\\$\SWPOTsmdiff$};
      \draw node[midway, above, draw=white, shape=rectangle] {$\Q$};
      \edge {T} {Y};
      \edge {X} {T};
      \edge {X} {Y};
    \end{influence-diagram}
    \caption{Synthetic data.}
    \label{fig: Q10 causal graph syn}
  \end{subfigure}
  \begin{subfigure}{0.5\textwidth}
    \centering
    \begin{influence-diagram}
      \node (help) [draw=none] {};      
      \node (T) [left = of help] {$T_A$};
      \node (Y) [right = of help] {$\RECO$};
      \node (X) [below = of help] {$\SWPOTsm$,\\$\SWPOTsmdiff$};
      \node (X_2) [above = of help] {$\VPD$};
      \draw node[midway, above, draw=white, shape=rectangle] {$\Q$};
      \edge {T} {Y};
      \edge {X} {T};
      \edge {X} {Y};
      \edge {T} {X_2};
      \edge {X_2} {Y};
    \end{influence-diagram}    
    \caption{Real data.}
    \label{fig: Q10 causal graph real}
  \end{subfigure}

  \caption{Assumed causal graphs for the estimation with the causal hybrid modeling approach in $\Q$ estimation.}
  \label{fig: Q10 causal graph}
\end{figure}

\paragraph{Causal graph of the $\Q$ model}\label{sec: Q10 causal graph}
The causal graph we assume for the $\Q$ model is shown in \cref{fig: Q10 causal graph}. The smooth potential radiation cycle given by $\SWPOTsm$ and $\SWPOTsmdiff$ represent seasonality and, thus, has a confounding effect on temperature $T_A$ and $\RECO$. For the real data, we add $\VPD$ to the graph, representing humidity and water availability.
This variable enters as a mediator in the graph as temperature affects evaporation and how much water the air can hold~\cite{Luo2006}. Furthermore, water availability also has a strong effect on respiration \cite{Chapin2013-ko}.
However, the temperature-sensitivity $\Q$ should only describe the immediate temperature effect~\cite{Luo2006}. We model the effects of water in the base respiration factor $\Rb$. Thus, assuming this graph, with our choices of variables, we estimate only the direct, immediate effect and not the one mediated through water or confounded by seasonality.

\subsection{\ce{CO2} Flux partitioning}
\subsubsection{Problem formulation}
Direct measurements of $GPP$ or $\RECO$ at the ecosystem level are difficult to obtain~\cite{Falge2003}. Alternatively, partitioning methods estimate these fluxes numerically from the measured $NEE$. Common approaches implement functional relationships based on physiology and estimate the fluxes using data-driven models~\cite{Reichstein2005, Moffat2007, Desai2008, Lasslop2010, Keenan2019}. Several hybrid-modeling approaches have recently been proposed modeling both fluxes with \acp{DNN} \cite{Tramontana2020, Trifunov2021, Zhan2022}. 

Separating a single signal into two additive signals is generally prone to equifinality issues. \cite{Tramontana2020} tried to break the symmetry between fluxes in the partition by enforcing different sets of explanatory environmental covariates for the two fluxes and applying a simple hybrid model. In particular, the authors combined \acp{DNN} with the light-use-efficiency model given by 
\begin{align}
    NEE = -LUE \cdot SW + \RECO, \label{eq:LUE}
\end{align}
where $\LUE$ models the linear efficiency of the incoming shortwaves $\SW$ on the resulting $\GPP$. In this form, $\GPP$ was modeled as the product of the incoming radiation and $\LUE$ parametrized by a \ac{DNN}. \cite{Zhan2022} showed that with different random initializations, this approach can lead to different resulting fluxes. The equifinality of the solution becomes particularly evident in extreme conditions. 
The authors can reduce variability through a multi-task learning approach. They introduce a second loss, forcing the network to learn to predict \ac{SIF} from the separated $GPP$ as both signals are known to be correlated under normal conditions.

\paragraph{Data}

As a proof of concept, we evaluate the proposed method on
synthetically generated data (see \cref{sec: linear flux partitioning}). We only used measured $NEE$ for the real data and applied the hybrid modeling approach site-wise per year.
For the data selection of real data from FLUXNET2015~\cite{Pastorello2020}, we closely followed~\cite{Tramontana2020} to compare our method to the neural network approach that imposes similar structural equations. 
We chose the same set of 36 different FLUXNET2015 sites (see \cref{sec: sites}) and used the same quality criterion to select site-years, i.e., years of a specific site. This implies that fitting is done year-wise per site, and only measured data is used. To have enough high-quality data, only site-years for the analysis are selected where at least 80\% of the meteorological data and 10\% of each daytime and nighttime $NEE$ were measured. As a target, similar to~\cite{Tramontana2020}, we use the $NEE$ obtained from the 50th percentile of the CUT method~\cite{Pastorello2020}. 
For comparison, we use the respective partitioned $\RECO$ and $\GPP$ fluxes obtained from the daytime~\cite{Lasslop2010} and nighttime~\cite{Reichstein2005} methods, already provided as part of the FLUXNET2015 dataset. Moreover, we compare the partitions to the results obtained with \acp{DNN} from~\cite{Tramontana2020}.

\paragraph{Applying \ac{DMLHM}}
We want to fit the following flux partitioning equation
\begin{align}
\NEE &= -\LUE(X)\cdot f(\SW) + \RECO(X,W),\label{eq:lte}
\end{align}
where $X$ and $W$ are sets of meteorological drivers and $f$ transforms the incoming radiation to allow for more flexible light-response curves, leading to a potentially non-linear light-use efficiency model. Here, $\RECO(\cdot)$ and $LUE(\cdot)$ represent $g(\cdot)$ and $\theta(\cdot)$ in the equivalent problem~\eqref{eq: problem}, respectively. This time, we use the estimator of $\RECO$ obtained from the first-stage estimators. As a proof of concept, we apply this method with $f$ being the identity function for linearly generated data over different noise levels (see \cref{sec: linear flux partitioning}).

For real data, the assumption of a linear relationship to $\SW$ is violated as $\GPP$ saturates with increasing light. We will thus first fit a transformation $f$ of the light curve before applying the \ac{DML} schema.
In order to find $f$, we finally fit $\alpha$ and $\beta$ in
\begin{align}
\NEE &= -\frac{\alpha \beta~\SW}{\alpha\SW + \beta} + \gamma.
\end{align}
with a moving window of $15$ days, we always transform the $5$ days in the center of the fitting interval. This procedure is motivated by the daytime flux partitioning method~\cite{Lasslop2010}, which estimates a parameterized rectangular hyperbola over moving windows to obtain $\GPP$. This heuristic allows us to find a flexible, smoothly changing light response curve. Other ways to obtain such a transformation can be envisaged. For the synthetic data, we use inputs according to how the data was generated, i.e., vapor pressure deficit $VPD$ and temperature $T_A$ for $X$ and the seasonal cycle of potential radiation for $W$. On the real data, we use day of the year $doy$, $VPD$, temperature $T_A$, and soil water content $SWC$ (for the sites where it is available) for $X$ and leave $W$ empty (For the assumed causal graphs, see \cref{sec: LUE causal graph}). We use gradient boosting regressors~\cite{Friedman2021}, an ensemble method of multiple shallow decision trees for all involved fitting steps.

\begin{figure}
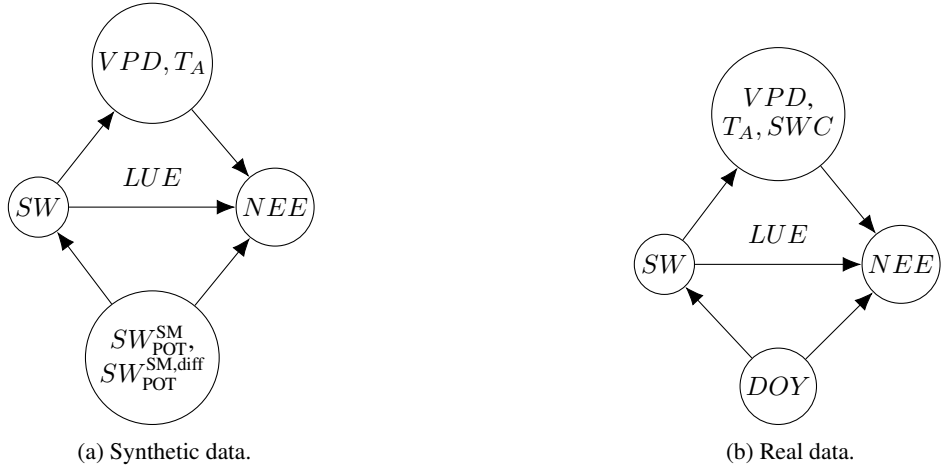

  \begin{subfigure}{0.5\textwidth}
    \centering
    \begin{influence-diagram}
      \node (help) [draw=none] {};      
      \node (T) [left = of help] {$\SW$};
      \node (Y) [right = of help] {$\NEE$};
      \node (X) [below = of help] {$\SWPOTsm$,\\$\SWPOTsmdiff$};
      \node (X_2) [above = of help] {$VPD, T_A$};
      \draw node[midway, above, draw=white, shape=rectangle] {$\LUE$};
      \edge {T} {Y};
      \edge {X} {T};
      \edge {X} {Y};
      \edge {T} {X_2};
      \edge {X_2} {Y};
    \end{influence-diagram}
    \caption{Synthetic data.}
    \label{fig: LUE causal graph syn}
  \end{subfigure}
  \begin{subfigure}{0.5\textwidth}
    \centering
    \begin{influence-diagram}
      \node (help) [draw=none] {};      
      \node (T) [left = of help] {$\SW$};
      \node (Y) [right = of help] {$\NEE$};
      \node (X) [below = of help] {$DOY$};
      \node (X_2) [above = of help] {$\VPD$,\\ $T_A, SWC$};
      \draw node[midway, above, draw=white, shape=rectangle] {$\LUE$};
      \edge {T} {Y};
      \edge {X} {T};
      \edge {X} {Y};
      \edge {T} {X_2};
      \edge {X_2} {Y};
    \end{influence-diagram}    
    \caption{Real data.}
    \label{fig: LUE causal graph real}
  \end{subfigure}
  \caption{Assumed causal graphs for the estimation with the causal hybrid modeling approach in flux partitioning.}
  \label{fig: LUE causal graph}
\end{figure}

\paragraph{Causal graph of the $\LUE$ model}\label{sec: LUE causal graph}
The causal graphs assumed for the $\LUE$ model are shown in \cref{fig: LUE causal graph}. As $\RECO$ is modeled similarly to the $\Q$ model, we keep the same variables modeling the seasonal cycle. In addition to that, we include $VPD$ and $T_A$, which were used to model $GPP$. The incoming radiation $\SW$ has an effect on the temperature as well as on water vapor~\cite{Luo2006}. Thus, both variables enter as mediators on the path to $\NEE$. For the real data, we use the day of the year $DOY$ to model the seasonality, which continues to be a confounder. In addition to the $VPD$ and $T_A$, we add soil water content, which also enters as a mediator when available. Consequently, we estimate $GPP$ as the direct effect of light on $NEE$, discounting the indirect effects through temperature, $VPD$, and $\SW$, which we allocate to $RECO$. Note that in this setup, these three variables are still entered as modifiers on the effect of light on $NEE$, affecting $GPP$.
\cref{tbl: inputs} summarizes the variables used for the different setups of the use cases.
\begin{table}
\centering
\caption{Summary of variables for the experiments.}
\label{tbl: inputs}
\begin{tabular}{llrrrr}
\toprule
Use case & Data & $Y$ & $T$ & $W$ &  $X$ \\
\midrule
$\Q$ model & Synthetic & $\log(\RECO)$ & $T_A$ & $\SWPOTsm$, $\SWPOTsmdiff$ & - \\
 \\
\cmidrule{2-6} 
 & Measured & $\log(\RECO)$ & $T_A$ & $\SWPOTsm$, $\SWPOTsmdiff$, $\VPD$& - \\
 & & & & &  \\
\midrule
\ce{CO2} Flux & Synthetic & $\NEE$ & $\SW$ & $\SWPOTsm$, $\SWPOTsmdiff$ & $\VPD$, $T_A$ \\ partitioning & & & & & \\
\cmidrule{2-6}
 & Measured & $\NEE$ & $\SW$ & $DOY$ & $\VPD$, $T_A, SWC$ \\
& & & & & \\
\bottomrule
\end{tabular}
\end{table}

\section{Results and Discussion}
\label{sec:results_PKL}
We show the applicability of our causal \ac{DMLHM} on two carbon flux modeling problems. We estimate the temperature sensitivity parameter in the $\Q$ model to showcase the robustness to regularization biases. We further illustrate the flexibility of the method to tackle the carbon flux partitioning problem.

\subsection{$\Q$ ecosystem respiration model.}
\subsubsection{Overall improved estimation capabilities.}
We simulated ecosystem respiration data from observations of FLUXNET. The true $\Q$ parameter was set to $1.5$. We sample $100$ datasets of varying sample sizes to see how the methods perform in different data regimes. We compare the \ac{GDHM} approach using \acp{DNN} to the proposed causal \ac{DMLHM} framework in two possible instantiations, either using \acp{RF} or \acp{DNN} as first-stage estimators. Experiments are run with and without applying dropout regularization and introducing $T_A$ as an additional predictor in base respiration.

\begin{figure}
  \begin{subfigure}{0.5\textwidth}
    \centering
     \includegraphics[width=1.0\textwidth]{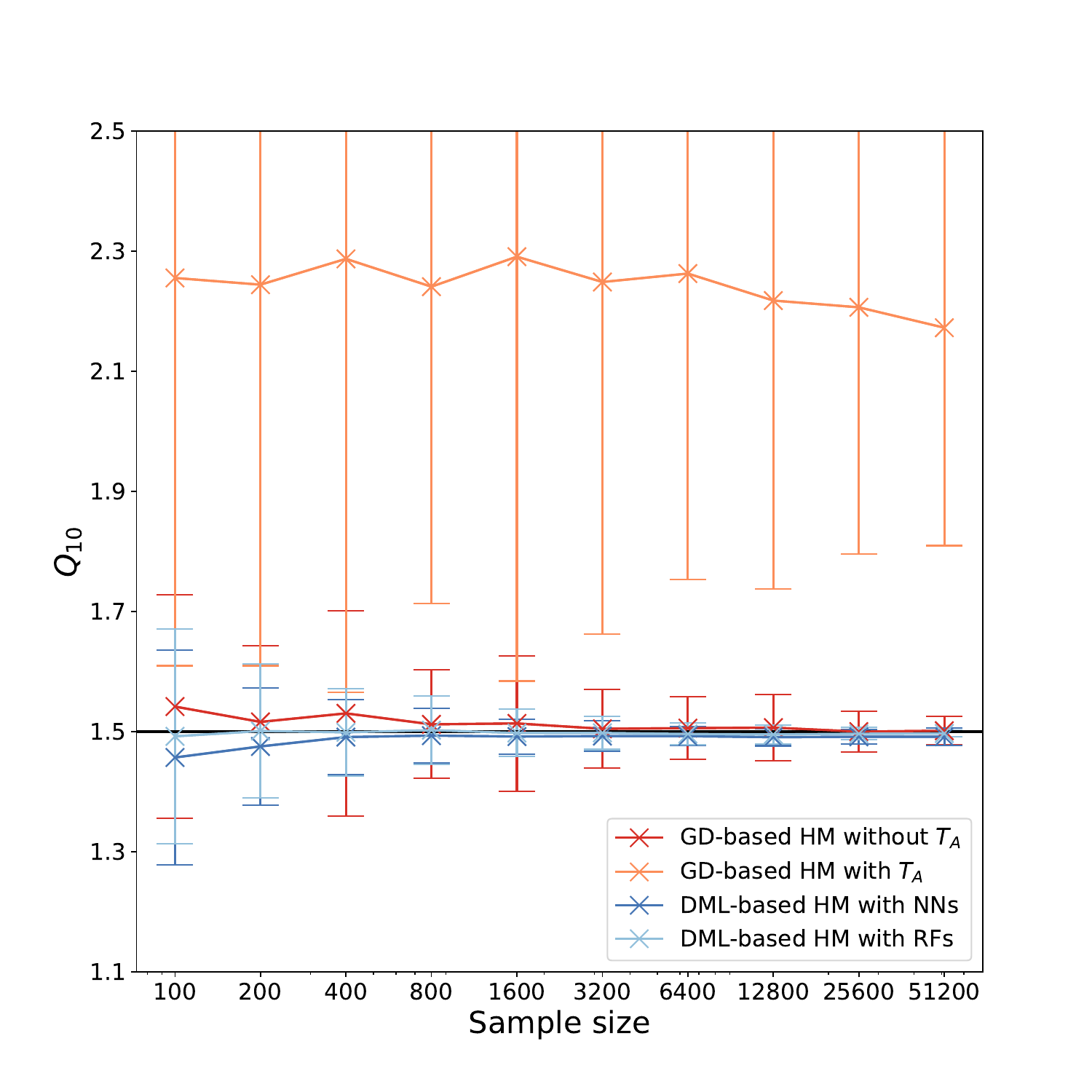}
    \caption{Without dropout.}
    \label{fig: Q10 without reg}
  \end{subfigure}
  \begin{subfigure}{0.5\textwidth}
    \centering
     \includegraphics[width=1.0\textwidth]{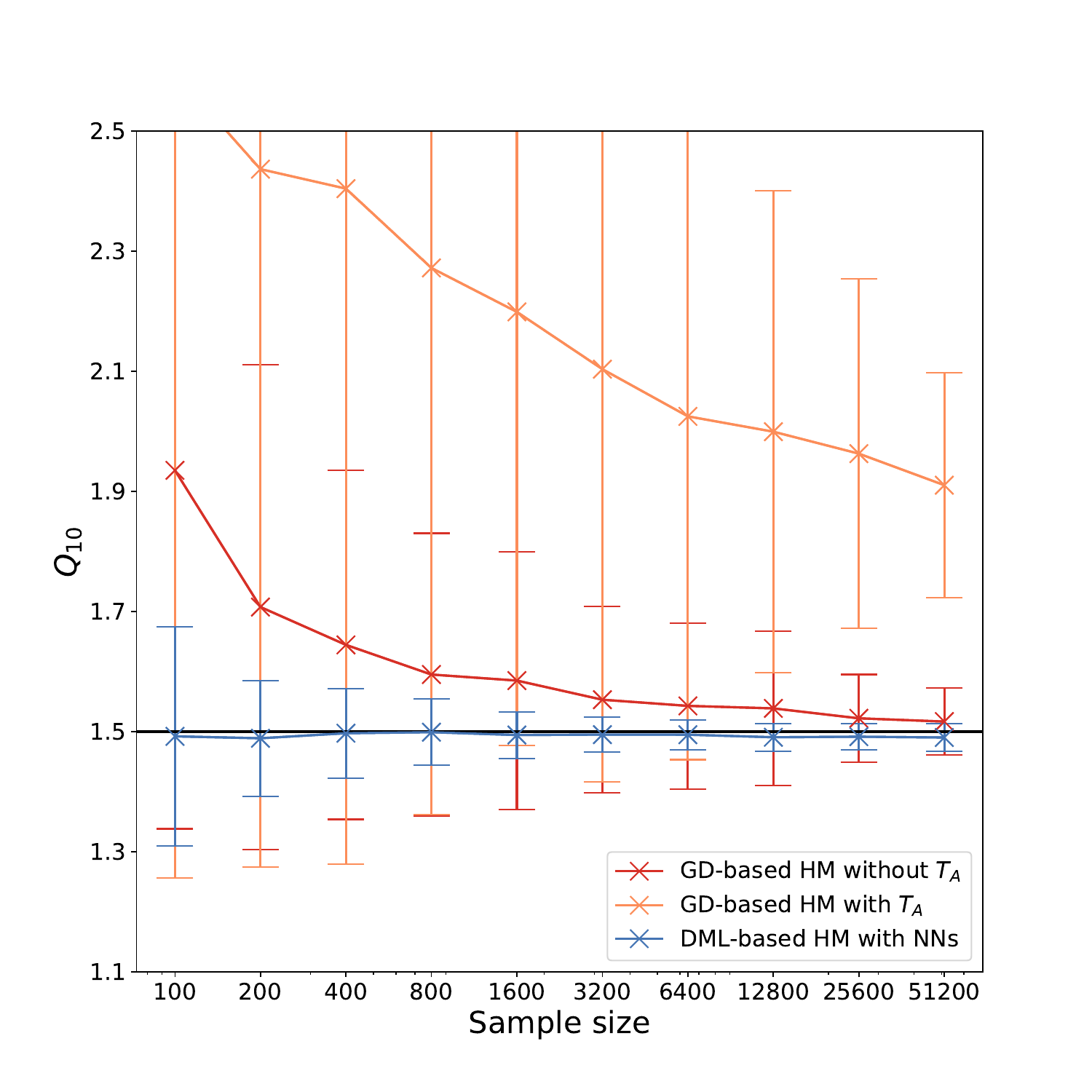}
    \caption{With dropout.}
    \label{fig: Q10 with reg}
  \end{subfigure}
  \caption{Simulation study for $\Q$ estimation with the \ac{GDHM} and the \ac{DMLHM} over 100 sampled datasets at different sample sizes. The plots show average and 95\% CI for the estimated $\Q$ for different methods without (a) and with (b) dropout applied as a regularizer in the \ac{DNN} regression models. The true $\Q$ parameter has a value of $1.5$. Introducing $T_A$ as a predictor in $\Rb$ leads to equifinality problems. Dropout as a regularizer introduces bias on the estimation of $\Q$ in the \ac{GDHM} case, while the causal hybrid modeling approach performs satisfactorily in the absence of equifinality.}
  \label{fig: Q10}
\end{figure}

The $\Q$ estimation results are shown in \cref{fig: Q10}. First, \cref{fig: Q10 without reg} shows the results where no dropout was applied to the \acp{DNN}. 
In this case, the estimates of the \ac{GDHM} approach, where $T_A$ is included as a predictor for $\Rb$, show values that are, on average, between $2.1$ and $2.3$ over all sample sizes. They show a substantial mismatch to the true value of $1.5$ and a wide spread at each sample size. This illustrates that equifinality expresses itself in the estimations as a wide range of values that hardly decreases with increasing sample size. We are not obtaining the full range of $\mathbb{R}>0$ values, which is by \eqref{eq:Q10} mathematically possible, but a range that is constraint alone by the initial $\Q$ value, the network's implicit biases and the first optimization steps of the gradient descent algorithm. This can make us mistake this for a valid inference of the method.
Instead, methods that exclude $T_A$ as a predictor find good estimators that converge with increasing data size. This is, in general, an encouraging result for all hybrid modeling approaches in this setup. Over the whole range, the \ac{GDHM} shows wider spreads than the \ac{DMLHM} approaches, which converge notably faster with increasing data size. At low data, they also have lower bias than the \ac{GDHM} approach. Remarkably, the random forest shows very little bias for solving this task over the whole data regime. Experiments corresponding to $\Q$ values of $1.25$ and $1.75$ (see \cref{sec: additional Q10 values}) exhibit minor variations in magnitude, proportional to the effect parameter. However, they consistently affirm the findings obtained for $\Q=1.5$.

These results showcase the data efficiency of the \ac{DML}-based approach. At the same time, it is currently computationally less efficient. 
The causal \ac{DMLHM} involves various fitting steps, which may seem uncomfortable compared to the usual end-to-end learning with \acp{DNN}. One may think of ways also to make \ac{DML} end-to-end possible. Here, one would apply \acp{NN} for all fitting steps and introduce a common loss over all optimization problems optimized with gradient descent. By weighting these losses adaptively, one can force this training to first fit the first stage estimators and then the treatment effect variable similar to what has been done in fitting PINNs respecting temporal and spatial causality \cite{wang2022respecting}. Efforts would need to be put into parallelizing the fitting of the first-stage estimators to make this approach computationally less costly.

\subsubsection{Robustness against regularization bias.}
Dropout is commonly used in deep learning for regularization\cite{Srivastava2014} or uncertainty quantification\cite{gal16}. \cref{fig: Q10 with reg} shows the $\Q$ estimations where dropout is applied to all \acp{DNN} of the \ac{GDHM} approach and the HM approach based on \ac{DML}. 
With dropout, the \ac{GDHM} approach has a more challenging time finding a good solution. It substantially overestimates the value of $\Q$ in the low data regime and only slowly gets more constrained and closer to the true value at the upper end of the used sample sizes. 
While the GD-based method got notably worse with the introduction of dropout, the \Ac{DML} shows robust results for the estimations over the full data range. On average, the $\Q$ estimations perform similarly to the experiments without dropout. In the low data regime, the bias in the estimation even decreased further.
When fitting the \ac{GDHM} with $T_A$, the regularization with dropout has a positive effect. The estimated values for $\Q$ are closer to the true value, and the spread reduces with more data points. The regularization through dropout restricts the space of solutions and reduces equifinality even though more data is necessary to overcome the stochasticity introduced through dropout. In \cref{sec: weight decay}, we show additional results with weight decay~\cite{Krogh1991}, another common regularization technique. As it yields qualitatively similar results (see \cref{fig: Q10 with weight decay}), we conclude that the presented findings are not only inherent to dropout.

In light of the results, \ac{DML}, in combination with dropout, can be effectively used for a full probabilistic assessment of hybrid models with inference on the parameter of interest and the non-parametric part, as dropout is also a common technique for obtaining uncertainty estimates for \acp{DNN} \cite{gal16}. While the \ac{GDHM} approach suffered from the application of dropout, the \ac{DML} approach was robust. Moreover, the technique further yields confidence bands for the approximately normally distributed estimators. By separating both estimations, we can obtain a distribution over the estimated $\Q$ and safely obtain uncertainty estimates for $\Rb$ using dropout.

\subsubsection{Results on real data}
\begin{figure}
    \centering
     \includegraphics[width=0.6\textwidth]{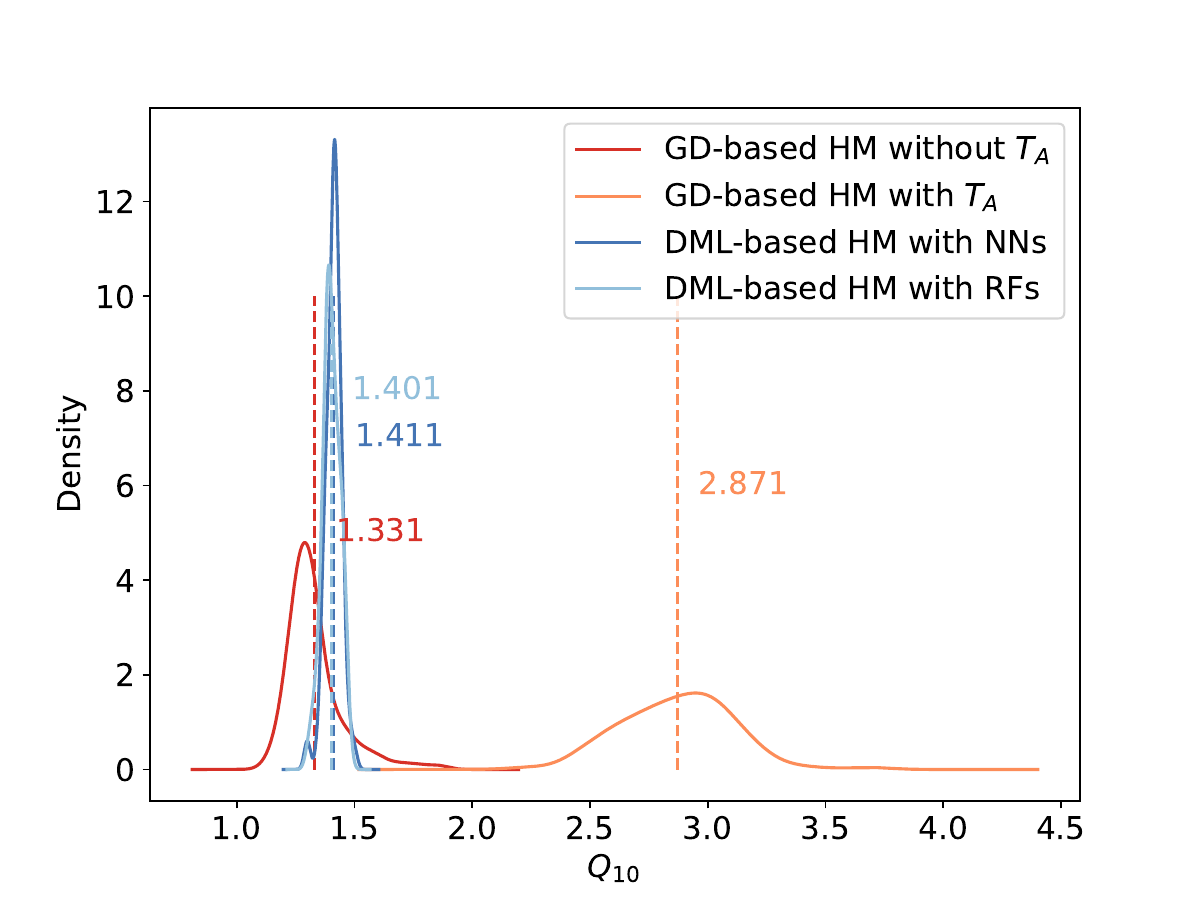}
   \caption{Estimation of $\Q$ on real data. Both \ac{DMLHM} find on average a $\Q$ value of $1.401$ and $1.411$ for \acp{RF} and \acp{NN}, respectively. This agrees with values from the literature that find a $\Q$ value around $1.41 \pm 0.1$~\cite{Mahecha2010}. The value for the \ac{GDHM} is lower at $1.331$ when leaving out $T_A$ as a predictor. With $T_A$, problems of equifinality show up again.}
  \label{fig: real Q10}
\end{figure}

As discussed in Section~\ref{sec:dataq10}, we obtain measured respiration data using nighttime $NEE$ measurements. 
We apply \ac{GDHM} and \ac{DMLHM} with \acp{DNN} and \acp{RF} without dropout to the data. 
We used the full dataset of over 100 different random seeds. The obtained distributions of $\Q$ are shown in~\cref{fig: real Q10}. The \ac{GDHM} approach finds a mean value of $1.322$, with a skewed distribution and estimated values ranging between $1$ and $2$. 
Including $T_A$ as a predictor in the GD-based approach, the values lie in a completely different range between $2.5$ and $3.5$, with the mean being $2.816$. 
The estimations based on \ac{DML} yield a mean of $1.407$ and $1.409$ for the \acp{RF} and \acp{DNN}, respectively, with similarly peaked distributions. The results of the \ac{DML} estimate agree fairly well with the results of~\cite{Mahecha2010} that after controlling for seasonal confounding, find that $\Q$ takes values around $1.41 \pm 0.1$ independently of mean-annual temperature and biome.

\subsection{\ce{CO2} flux partitioning}
We apply the causal \ac{DMLHM} to the problem of carbon flux partitioning as defined in \eqref{eq:nee}. 
In this scenario, we model the effect as a heterogeneous treatment effect, a function of other predictors, parametrized with an \ac{ML} model. We use gradient boosting estimators for all three estimators involved. Moreover, we show that the plug-in estimator for $\RECO$ 
obtained by combining the first-stage estimators yields useful values without the need for an additional refit.

\subsubsection{Consistent flux partitioning}
We use vapor pressure deficit $VPD$, air temperature $T_A$, and day of the year (for seasonality) as drivers over all sites. 
Where available, we also included soil water content.
Since we do not have access to the real partial fluxes, we compare the retrieved fluxes to the ones obtained by the \ac{DNN} approach described in~\cite{Tramontana2020} and by the established daytime and nighttime methods~\cite{Lasslop2010, Reichstein2005}.
The daytime and nighttime methods are assumed to capture a simple cycle depending on a few meteorological drivers. New methods may deviate but should show a similar pattern overall.
For the partitioned fluxes of two methods $(x_i)_{i=1}^N$ and $(y_i)_{i=1}^N$, we compute the $R^2$, the \ac{RMSE}, given by
$\sqrt{\frac{\sum_{i=1}^N(x_i-y_i)^2}{N}}$, and the bias as the difference between the sample means $\Bar{x}$ and $\Bar{y}$. 
The results are reported in~\cref{tbl:results_partitioning}. 
\begin{table}
\centering
\caption{
Cross consistency in terms of $R^2$, $RMSE$ and bias of retrieved $GPP$, $RECO$ and estimated $NEE$ between the established daytime (DT)~\cite{Lasslop2010} and nighttime (NT)~\cite{Reichstein2005} methods and the \ac{GDHM} with neural networks (NN)~\cite{Tramontana2020} and \ac{DMLHM} (DML), proposed in this work. The reported statistics are median and in brackets 0.25/0.75 quantiles over all site-years.
}
\label{tbl:results_partitioning}
\begin{tabular}{llLLL}
\toprule
 Flux & Methods &    R^{{2}*} &   RMSE^* (\frac{\mu \mol \ce{CO2}}{m^{2}s}) & Bias (\frac{\mu \mol \ce{CO2}}{m^{2}s}) \\
\midrule
RECO & DT  vs. DML & 0.62 (0.41/0.74) & 1.18 (0.75/1.46) & \phantom{-}0.00 (-0.20/0.14) \\
     & DT  vs. NN & 0.69 (0.50/0.81) & 0.98 (0.70/1.29) & \phantom{-}0.02 (-0.12/0.18) \\
     & NT  vs. DML & 0.74 (0.50/0.83) & 0.89 (0.57/1.15) & \phantom{-}0.00 (-0.11/0.10) \\
     & NT  vs. NN & 0.85 (0.65/0.92) & 0.68 (0.47/0.84) & \phantom{-}0.07 (-0.02/0.16) \\
     & DT  vs. NT  & 0.73 (0.63/0.83) & 0.95 (0.64/1.21) & \phantom{-}0.00 (-0.22/0.16) \\
     & NN vs. DML & 0.63 (0.34/0.77) & 0.99 (0.66/1.24) & -0.07 (-0.22/0.10) \\
\midrule
GPP & DT  vs.  DML & 0.96 (0.93/0.97) & 1.25 (0.74/1.49) & \phantom{-}0.00 (-0.16/0.11) \\
    & DT  vs.  NN & 0.96 (0.93/0.97) & 1.22 (0.76/1.52) & \phantom{-}0.04 (-0.04/0.17) \\
    & NT  vs.  DML & 0.90 (0.84/0.92) & 1.97 (1.16/2.47) &  -0.02 (-0.13/0.10) \\
    & NT  vs.  NN & 0.93 (0.89/0.95) & 1.53 (0.90/2.02) & \phantom{-}0.07 (-0.02/0.18) \\
    & DT  vs.  NT  & 0.89 (0.82/0.92) & 1.85 (1.20/2.42) & \phantom{-}0.02 (-0.16/0.13) \\
    & NN vs.  DML & 0.95 (0.92/0.97) & 1.32 (0.71/1.61) & -0.08 (-0.23/0.08) \\
\midrule
NEE & DT  vs.  DML & 0.95 (0.93/0.97) & 1.07 (0.71/1.29) & -0.02 (-0.11/0.07) \\
    & DT  vs.  NN & 0.94 (0.91/0.96) & 1.13 (0.76/1.36) & -0.03 (-0.12/0.03) \\
    & NT$^{\text{*}}$  vs.  DML & 0.87 (0.81/0.89) & 1.92 (1.15/2.36) & \phantom{-}0.01 (-0.02/0.06) \\
    & NT$^{\text{*}}$  vs.  NN & 0.93 (0.90/0.94) & 1.29 (0.79/1.82) & \phantom{-}0.00 (-0.01/0.01) \\
    & DT  vs.  NT$^{\text{*}}$  & 0.86 (0.79/0.90) & 1.68 (1.12/2.25) & -0.03 (-0.12/0.03) \\
    & NN vs.  DML & 0.94 (0.91/0.96) & 1.27 (0.77/1.52) & \phantom{-}0.01 (-0.02/0.05) \\
\bottomrule
\end{tabular}
\\
$^*$\textnormal{The NT NEE value corresponds exactly to the measured NEE value.}
\end{table}

Overall the consistency of the method based on \ac{DML} lies in a similar range of values to the \ac{DNN} approach~\cite{Tramontana2020} 
when compared to the daytime and nighttime methods. 
The estimated data uncertainty of the used $\NEE$ measurements is $1.53 \frac{\mu \mol \ce{CO2}}{m^{2}s}$. For almost all compared fluxes, our method lies under this threshold in terms of \ac{RMSE}. Only for the $\GPP$ and $\NEE$ of the nighttime method, the values lie on average slightly above with $1.97 \frac{\mu \mol \ce{CO2}}{m^{2}s}$ and $1.92 \frac{\mu \mol \ce{CO2}}{m^{2}s}$, respectively.
The nighttime method fits respiration overnight and obtains $\GPP$ as the residuals between the estimated $\RECO$ and measured $\NEE$. Thus, by construction, the $\NEE$ of the nighttime method corresponds to the measured $\NEE$. Hence, both $\NEE$ and $\GPP$ of the nighttime method are higher in noise, and thus, a higher \ac{RMSE} of our method is expected. 
When comparing the bias between methods, the causal \ac{DMLHM} shows a slightly smaller bias compared to both standard methods than these methods between them in almost all cases. Furthermore, it lies in a similar range to the \ac{GDHM}.

Overall, our method shows higher similarity to the daytime method, which is expected due to the fitting of the rectangular hyperbola in the first step. The retrieved $\GPP$ is similar to the daytime method as the \ac{DNN} approach, and the obtained $\NEE$ is even closer. At the same time, the obtained $\RECO$ shows a larger deviation even to the daytime method. This is because we used the plugin-in estimator for $\RECO$ obtained from the first-stage \ac{DML} estimators. 

We could obtain a more sophisticated estimator by refitting another model on the residuals, as done in the case of the $\Q$ model, where we could also employ $\SW$ as a predictor without experiencing equifinality. It would even allow using the previously estimated $\GPP$ as a predictor of $\RECO$.
As an additional proof of concept, we apply the method to synthetic data with different levels of heteroscedastic noise. The method finds robust estimates even to high levels of noise. The results can be found in \cref{sec: linear flux partitioning}.

\subsubsection{Learned functionalities}
\begin{figure}
    \centering
     \includegraphics[width=0.9\textwidth]{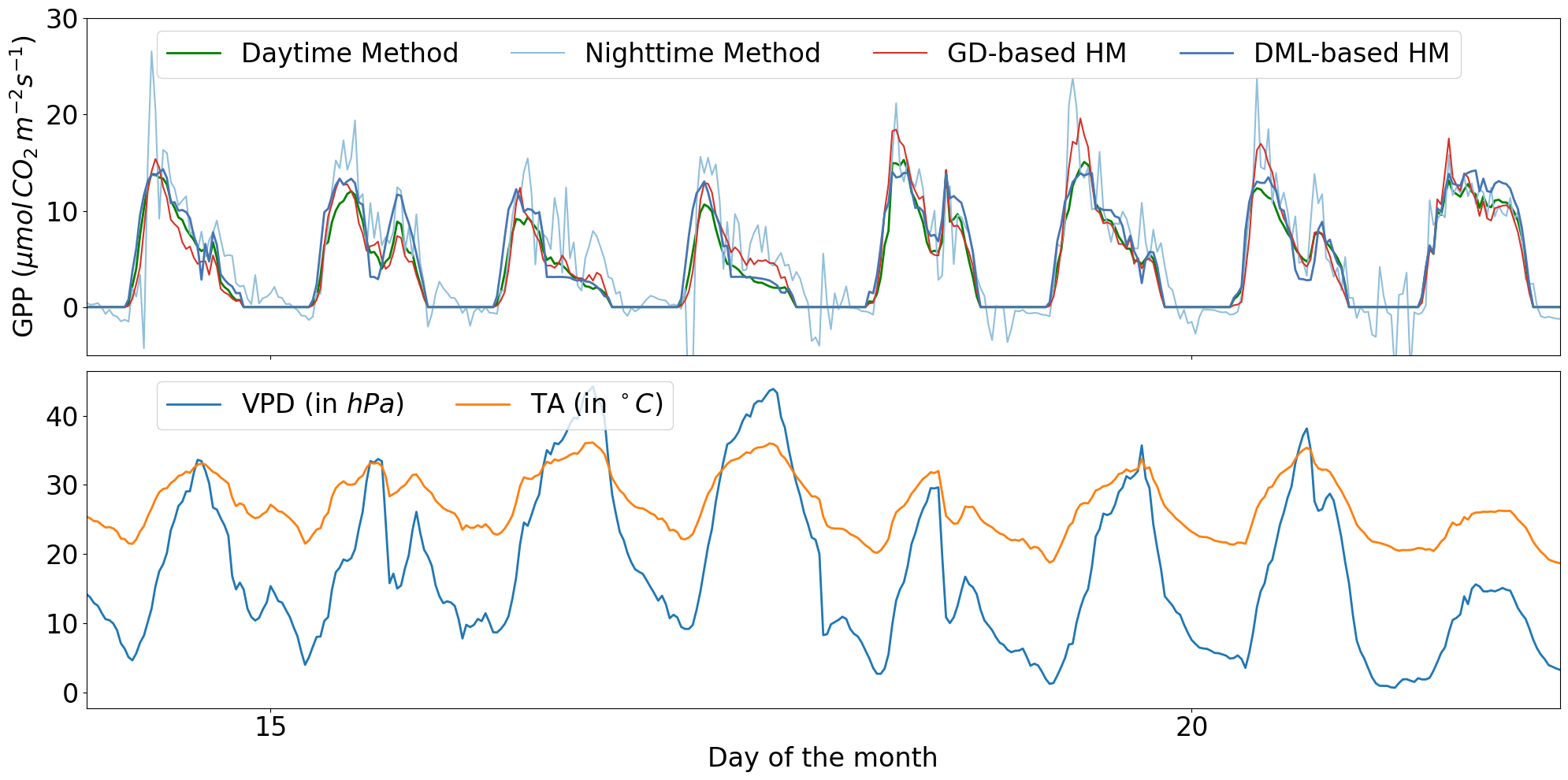}
     \caption{Retrieved $\GPP$ flux of daytime method, nighttime method and \ac{DMLHM} in July 2006 in France Le-Bray. The \ac{DMLHM} retrieved a similar flux to the daytime method that decreases with the increase of $VPD$.}
    \label{fig: GPP_FR-LBr}
\end{figure}

\begin{figure}
    \centering
     \includegraphics[width=0.9\textwidth]{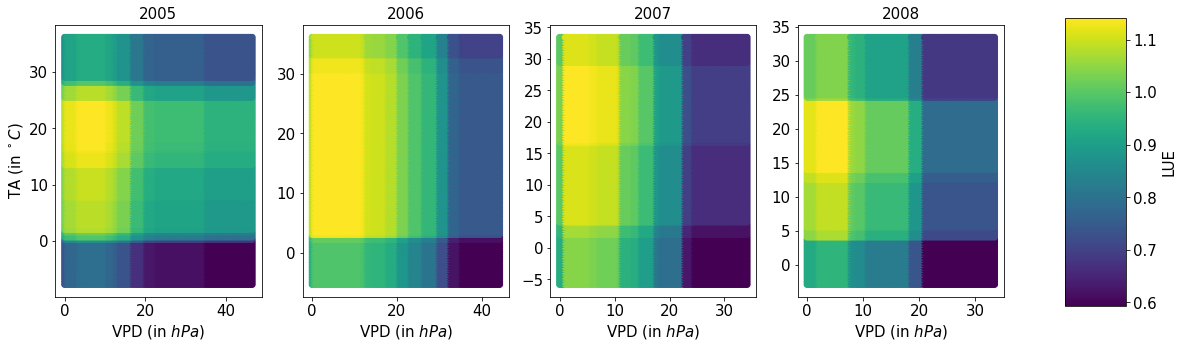}
     \caption{Functional behavior of the learned LUE in the years 2005 to 2008 over $VPD$ and $TA$. The LUE shows a consistent functionality over the different years where an increase in $VPD$, which marks lower water availability, reduces productivity. This is also consistent with the functionality that the daytime method implements parametrically.}
    \label{fig: functional_rel_VPD_TA_FR-LBr}
\end{figure}

The consistency tables served as a sanity check that the methods produce reasonable estimations that contain similar trends over the day and year. 
The next questions are: Where do they produce similar outputs? When do the outputs differ? 
For this, we compare the retrieved fluxes on two different sites. 
In~\cref{fig: GPP_FR-LBr}, we see the retrieved $\GPP$ flux over a few days in July 2006 in France Le Bray. %
We compare the \ac{DMLHM} to the \ac{GDHM}, daytime and nighttime methods. The retrieved $\GPP$ of the daytime and hybrid modeling methods show similar patterns. High $\VPD$, which marks low water availability, reduces productivity. The daytime method implements this functionality parametrically. The $\LUE$ function of the \ac{DMLHM} approach learned a similar functionality that decreases with increasing $\VPD$ and has preferred temperatures roughly between $15^\circ C$ and $30^\circ C$ (see \cref{fig: functional_rel_VPD_TA_FR-LBr}). It is consistent over the four consecutive years the method was applied to at this site. This demonstrates that the causal hybrid modeling approach can learn a similar functional relationship as the parametric daytime method in a non-parametric way. The nighttime method shows a noisier but qualitatively similar pattern.

\begin{figure}
    \centering
     \includegraphics[width=0.9\textwidth]{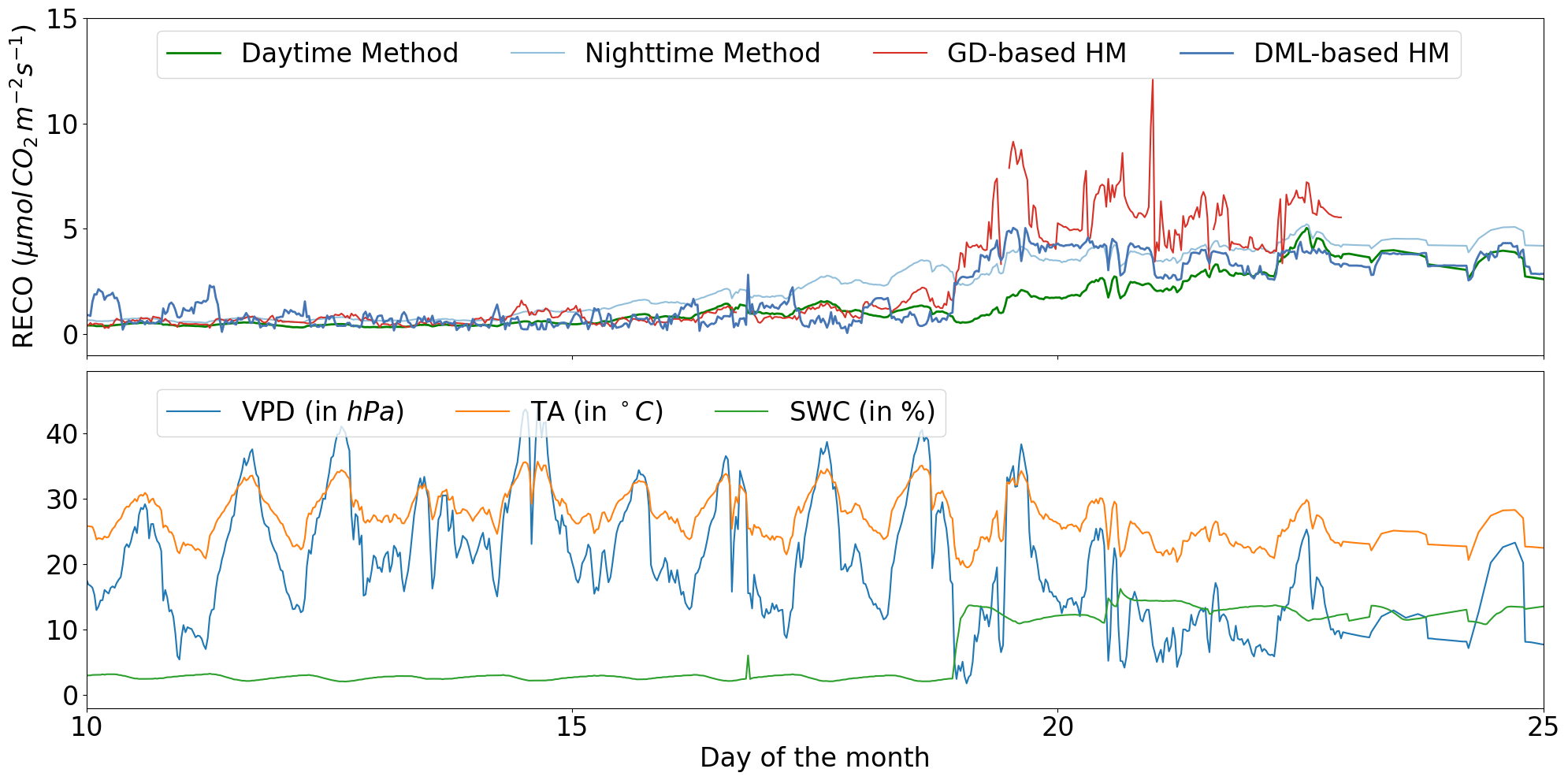}
     \caption{Retrieved $\RECO$ flux of daytime, nighttime, and both hybrid modeling methods in July 2010 in Santa Rita in the US. The daytime and nighttime methods show slow adaption to the change in dynamics caused by a rain pulse event that followed a long drought. Both hybrid modeling approaches can retrieve the expected immediate increase in respiration. The estimates of the \ac{GDHM} are lower and less noisy.}
    \label{fig: RECO_US-SRG}
\end{figure}

To highlight the differences between the methods, we look at a grassland site in Santa Rita (US)~\cite{Scott2015}. \cref{fig: RECO_US-SRG} shows the estimated $\RECO$ over few days in July 2010. The selected time window was preceded by two months without rain, leading to low soil water content and, in turn, reduced respiration activity~\cite{Chapin2013-ko}.
During the shown period, a rain event leads to a sudden increase in soil water content. 
Such an event is expected to lead to a sudden increase in respiration as it stimulates microbial activity~\cite{Chapin2013-ko}. We find that the daytime and nighttime methods cannot capture this sudden behavior as their estimation is based on window fitting and cannot detect sudden changes in dynamics. While $\RECO$ estimated with the nighttime method increases even before the event, the daytime method yields slowly increasing respiration flux shortly after the event. Instead, the fluxes estimated with the non-parametric hybrid modeling approaches show an increase right at the event's time, demonstrating that they can adapt to sudden changes in dynamics. A difference between both hybrid modeling approaches shows that the \ac{GDHM} estimates a stronger respiration pulse but yields a noisier estimate from the onset of the event. 

Our approach offers unique advantages. While traditional daytime and nighttime methods are fully interpretable, they struggle to capture rapid dynamic changes due to their parametric nature. On the other hand, the end-to-end 
GD-based methods, such as the approach by~\cite{Tramontana2020}, may lack interpretability due to non-identifiability or implicit functional constraints, relying on assumptions with unclear implications.
In contrast, our causal interpretation-based approach offers a middle ground, providing reasonable estimates of fluxes while maintaining interpretability as it is grounded in causal assumptions. By identifying GPP as the causal effect of light on NEE, our method offers a clear and meaningful interpretation of the flux partitioning process. While it may not match the predictive performance and flexibility of pure deep learning, it offers a valuable alternative by combining interpretability with reasonable estimation accuracy.

The analysis we carried out merely serves as a proof of concept toward a causally meaningful flux partitioning method. To maintain comparability, we ran the experiments on the same sites and years with similar quality filters as~\cite{Tramontana2020}. For both \ac{DMLHM} as well as the  \ac{GDHM} approach with \acp{NN}, further research is necessary before they can be employed at scale in the data processing pipelines of FLUXNET sites. In particular, this would require a comprehensive analysis of the performance over all FLUXNET sites to disentangle the effects of geographical region, climate, vegetation, data quality, and data availability on the consistency of new flux partitioning methods. This should ideally be accompanied by simulations of sets of land surface models tailored for different land cover types to benchmark the adaptability of data-driven methods. This is beyond the scope of this work, which aims at introducing a causal approach to hybrid modeling.
As for today, a benchmarking set and standardized evaluation pipeline are not available but could become key in the future when more data-driven flux partitioning models are developed. 
Understanding how these local factors influence the data-driven methods is crucial as the flux partitioning products serve as ground truth for downstream tasks such as upscaling from the site level to global fluxes as aimed for in the FLUXCOM project~\cite{Jung2020}.

\section{Conclusions}
Machine learning is becoming a complementary tool to enhance scientific research and discovery in all fields of science.
Its limitations are evident: lack of transparency and interpretability, weak generalizability to unseen data, and violation of governing laws.
Hybrid modeling aims to incorporate scientific knowledge to overcome these limitations. However, this alone is insufficient to obtain the interpretability we hope for. Spurious links between variables can lead to equifinality: many models describe the data similarly well. Therefore, we must also teach these hybrid models what seems evident to us: correlation is not causation. And it is causation that we want.

In this paper, we propose a first step in this direction. We split the fitting of hybrid modeling involving treatment effects into subsequent steps, where we first estimated the causal effect with \ac{DML} and then estimated the remaining of the model.
By separating different estimation steps and being explicit about the underlying causal graph and the causal effect, 
we were able to obtain a well-defined problem that, 
originally was ill-posed and, in practice, suffering from equifinality.
We applied this technique to two problems of carbon flux estimation, namely, $\Q$ estimation in ecosystem respiration and carbon flux partitioning. 
We demonstrated the superiority of \ac{DML} in retrieving parameters describing causal effects over end-to-end estimations with usual hybrid modeling approaches using \acp{DNN}. 
The estimation is shown to be efficient and robust and effectively reduces bias 
through regularization techniques such as dropout and weight decay. 
On real data, it could retrieve a value for $\Q$ consistent with the literature.
We further showed the flexibility of the method by transforming the treatment and fitting a heterogeneous treatment effect of the $LUE$ model for carbon flux partitioning as a non-parametric function.
The retrieved fluxes were consistent with the ones of established methods, showed reasonable functional dependencies, and could improve on known limitations stemming from the window fitting of these methods.

We note that to apply the method effectively, assuming a causal graph and being explicit about the causal relationships of the involved variables is essential. This also includes thinking about unobserved confounders, mediators, and correlations between variables. We believe that this should be a general best practice. Our method encourages machine learners and practitioners to do so. %
A remaining problem is that even though we could show that it has broader applicability than the standard semi-linear regression problem, its relevance is still limited to hybrid models of a particular form containing parameters or non-parametric functions describing causal effects.

Integrating causality with hybrid modeling is crucial for achieving more interpretable and reliable outcomes in knowledge-driven machine learning. Our work has showcased this integration in two important problems in ecology through the application of causal effect estimation. Our causal hybrid modeling framework holds promise for enhancing interpretability and causal inference across diverse scientific fields that demand more insightful machine learning models. Looking ahead, we encourage further exploration and integration of causality concepts within hybrid modeling techniques.
\section*{Acknowledgments}

This work received support from the European Research Council (ERC) under the ERC Synergy Grant USMILE (grant agreement 855187). We express our gratitude to Gianluca Tramontana for generously providing his data and patiently answering all our queries.

\bibliographystyle{IEEEtran}
\bibliography{main}

\begin{thebibliography}{10}
\providecommand{\url}[1]{#1}
\csname url@samestyle\endcsname
\providecommand{\newblock}{\relax}
\providecommand{\bibinfo}[2]{#2}
\providecommand{\BIBentrySTDinterwordspacing}{\spaceskip=0pt\relax}
\providecommand{\BIBentryALTinterwordstretchfactor}{4}
\providecommand{\BIBentryALTinterwordspacing}{\spaceskip=\fontdimen2\font plus
\BIBentryALTinterwordstretchfactor\fontdimen3\font minus \fontdimen4\font\relax}
\providecommand{\BIBforeignlanguage}[2]{{%
\expandafter\ifx\csname l@#1\endcsname\relax
\typeout{** WARNING: IEEEtran.bst: No hyphenation pattern has been}%
\typeout{** loaded for the language `#1'. Using the pattern for}%
\typeout{** the default language instead.}%
\else
\language=\csname l@#1\endcsname
\fi
#2}}
\providecommand{\BIBdecl}{\relax}
\BIBdecl

\bibitem{kirillov2023segment}
A.~Kirillov, E.~Mintun, N.~Ravi, H.~Mao, C.~Rolland, L.~Gustafson, T.~Xiao, S.~Whitehead, A.~C. Berg, W.-Y. Lo, P.~Doll{\'{a}}r, and R.~Girshick, ``Segment anything,'' 2023.

\bibitem{brown2020language}
T.~B. Brown, B.~Mann, N.~Ryder, M.~Subbiah, J.~Kaplan, P.~Dhariwal, A.~Neelakantan, P.~Shyam, G.~Sastry, A.~Askell, S.~Agarwal, A.~Herbert-Voss, G.~Krueger, T.~Henighan, R.~Child, A.~Ramesh, D.~M. Ziegler, J.~Wu, C.~Winter, C.~Hesse, M.~Chen, E.~Sigler, M.~Litwin, S.~Gray, B.~Chess, J.~Clark, C.~Berner, S.~McCandlish, A.~Radford, I.~Sutskever, and D.~Amodei, ``Language models are few-shot learners,'' 2020.

\bibitem{zhang2022pushing}
Y.~Zhang, J.~Qin, D.~S. Park, W.~Han, C.-C. Chiu, R.~Pang, Q.~V. Le, and Y.~Wu, ``Pushing the limits of semi-supervised learning for automatic speech recognition,'' 2022.

\bibitem{Halevy09}
\BIBentryALTinterwordspacing
A.~Halevy, P.~Norvig, and F.~Pereira, ``The unreasonable effectiveness of data,'' \emph{IEEE Intelligent Systems}, vol.~24, no.~2, pp. 8--12, Mar. 2009. [Online]. Available: \url{http://dx.doi.org/10.1109/MIS.2009.36}
\BIBentrySTDinterwordspacing

\bibitem{Lipton18}
\BIBentryALTinterwordspacing
Z.~C. Lipton, ``The mythos of model interpretability,'' \emph{Queue}, vol.~16, no.~3, pp. 30:31--30:57, Jun. 2018. [Online]. Available: \url{http://doi.acm.org/10.1145/3236386.3241340}
\BIBentrySTDinterwordspacing

\bibitem{kump2013earth}
L.~R. Kump, J.~F. Kasting, and R.~G. Crane, \emph{The Earth System}, 3rd~ed.\hskip 1em plus 0.5em minus 0.4em\relax Pearson, 2013.

\bibitem{O'Neill2016}
\BIBentryALTinterwordspacing
B.~C. O'Neill, C.~Tebaldi, D.~P. van Vuuren, V.~Eyring, P.~Friedlingstein, G.~Hurtt, R.~Knutti, E.~Kriegler, J.-F. Lamarque, J.~Lowe, G.~A. Meehl, R.~Moss, K.~Riahi, and B.~M. Sanderson, ``The scenario model intercomparison project ({ScenarioMIP}) for {CMIP6},'' \emph{Geoscientific Model Development}, vol.~9, no.~9, pp. 3461--3482, 2016. [Online]. Available: \url{https://gmd.copernicus.org/articles/9/3461/2016/}
\BIBentrySTDinterwordspacing

\bibitem{Eyring2016}
\BIBentryALTinterwordspacing
V.~Eyring, S.~Bony, G.~A. Meehl, C.~A. Senior, B.~Stevens, R.~J. Stouffer, and K.~E. Taylor, ``Overview of the coupled model intercomparison project phase 6 ({CMIP6}) experimental design and organization,'' \emph{Geoscientific Model Development}, vol.~9, no.~5, pp. 1937--1958, 2016. [Online]. Available: \url{https://gmd.copernicus.org/articles/9/1937/2016/}
\BIBentrySTDinterwordspacing

\bibitem{Myers2021}
T.~A. {Myers}, R.~C. {Scott}, M.~D. {Zelinka}, S.~A. {Klein}, J.~R. {Norris}, and P.~M. {Caldwell}, ``{Observational constraints on low cloud feedback reduce uncertainty of climate sensitivity},'' \emph{Nature Climate Change}, vol.~11, no.~6, pp. 501--507, Jan. 2021.

\bibitem{Hewitt2020}
\BIBentryALTinterwordspacing
H.~T. Hewitt, M.~Roberts, P.~Mathiot, A.~Biastoch, E.~Blockley, E.~P. Chassignet, B.~Fox-Kemper, P.~Hyder, D.~P. Marshall, E.~Popova, A.-M. Treguier, L.~Zanna, A.~Yool, Y.~Yu, R.~Beadling, M.~Bell, T.~Kuhlbrodt, T.~Arsouze, A.~Bellucci, F.~Castruccio, B.~Gan, D.~Putrasahan, C.~D. Roberts, L.~Van~Roekel, and Q.~Zhang, ``Resolving and parameterising the ocean mesoscale in earth system models,'' \emph{Current Climate Change Reports}, vol.~6, no.~4, pp. 137--152, Dec 2020. [Online]. Available: \url{https://doi.org/10.1007/s40641-020-00164-w}
\BIBentrySTDinterwordspacing

\bibitem{YUAN2022108920}
\BIBentryALTinterwordspacing
K.~Yuan, Q.~Zhu, W.~J. Riley, F.~Li, and H.~Wu, ``Understanding and reducing the uncertainties of land surface energy flux partitioning within {CMIP6} land models,'' \emph{Agricultural and Forest Meteorology}, vol. 319, p. 108920, 2022. [Online]. Available: \url{https://www.sciencedirect.com/science/article/pii/S0168192322001137}
\BIBentrySTDinterwordspacing

\bibitem{Arora2020}
\BIBentryALTinterwordspacing
V.~K. Arora, A.~Katavouta, R.~G. Williams, C.~D. Jones, V.~Brovkin, P.~Friedlingstein, J.~Schwinger, L.~Bopp, O.~Boucher, P.~Cadule, M.~A. Chamberlain, J.~R. Christian, C.~Delire, R.~A. Fisher, T.~Hajima, T.~Ilyina, E.~Joetzjer, M.~Kawamiya, C.~D. Koven, J.~P. Krasting, R.~M. Law, D.~M. Lawrence, A.~Lenton, K.~Lindsay, J.~Pongratz, T.~Raddatz, R.~S\'ef\'erian, K.~Tachiiri, J.~F. Tjiputra, A.~Wiltshire, T.~Wu, and T.~Ziehn, ``Carbon--concentration and carbon--climate feedbacks in {CMIP6} models and their comparison to {CMIP5} models,'' \emph{Biogeosciences}, vol.~17, no.~16, pp. 4173--4222, 2020. [Online]. Available: \url{https://bg.copernicus.org/articles/17/4173/2020/}
\BIBentrySTDinterwordspacing

\bibitem{Zhu2014}
\BIBentryALTinterwordspacing
Q.~Zhu and Q.~Zhuang, ``Parameterization and sensitivity analysis of a process-based terrestrial ecosystem model using adjoint method,'' \emph{Journal of Advances in Modeling Earth Systems}, vol.~6, no.~2, pp. 315--331, 2014. [Online]. Available: \url{https://agupubs.onlinelibrary.wiley.com/doi/abs/10.1002/2013MS000241}
\BIBentrySTDinterwordspacing

\bibitem{reichstein19nat}
M.~Reichstein, G.~Camps-Valls, B.~Stevens, J.~Denzler, N.~Carvalhais, M.~Jung, and Prabhat, ``Deep learning and process understanding for data-driven { E } arth system science,'' \emph{Nature}, vol. 566, pp. 195--204, Feb 2019.

\bibitem{Camps-Valls2021}
G.~Camps-Valls, D.~Tuia, X.~X. Zhu, and M.~Reichstein, \emph{Bibliography}.\hskip 1em plus 0.5em minus 0.4em\relax John Wiley \& Sons, Ltd, 2021, pp. 331--400.

\bibitem{CampsValls09wiley}
G.~Camps-Valls and L.~Bruzzone, \emph{Kernel methods for Remote Sensing Data Analysis}.\hskip 1em plus 0.5em minus 0.4em\relax UK: Wiley \& Sons, Dec 2009.

\bibitem{Tramontana16bg}
\BIBentryALTinterwordspacing
G.~Tramontana, M.~Jung, G.~Camps-Valls, K.~Ichii, B.~Raduly, M.~Reichstein, C.~R. Schwalm, M.~A. Arain, A.~Cescatti, G.~Kiely, L.~Merbold, P.~Serrano-Ortiz, S.~Sickert, S.~Wolf, and D.~Papale, ``Predicting carbon dioxide and energy fluxes across global {FLUXNET} sites with regression algorithms,'' \emph{Biogeosciences Discussions}, vol. 2016, pp. 1--33, 2016. [Online]. Available: \url{http://www.biogeosciences-discuss.net/bg-2015-661/}
\BIBentrySTDinterwordspacing

\bibitem{CampsValls21wiley}
\BIBentryALTinterwordspacing
G.~Camps-Valls, D.~Tuia, X.~X. Zhu, and M.~R. (Editors), \emph{Deep learning for the {E}arth Sciences: {A} comprehensive approach to remote sensing, climate science and geosciences}.\hskip 1em plus 0.5em minus 0.4em\relax Wiley \& Sons, 2021. [Online]. Available: \url{https://github.com/DL4ES}
\BIBentrySTDinterwordspacing

\bibitem{Rudin2019Why}
C.~Rudin and J.~Radin, ``Why are we using black box models in {AI} when we don\textquoteright{}t need to? {A} lesson from an explainable {AI} competition,'' \emph{Harvard Data Science Review}, vol.~1, no.~2, nov 22 2019, https://hdsr.mitpress.mit.edu/pub/f9kuryi8.

\bibitem{QuioneroCandela2009}
J.~Quionero-Candela, M.~Sugiyama, A.~Schwaighofer, and N.~D. Lawrence, ``Dataset shift in machine learning,'' 2009.

\bibitem{6278037}
M.~Sugiyama and M.~Kawanabe, \emph{Learning Under Covariate Shift}, 2012, pp. 19--19.

\bibitem{marcus2018deep}
G.~Marcus, ``Deep learning: {A} critical appraisal,'' \emph{arXiv preprint arXiv:1801.00631}, 2018.

\bibitem{IPCC_WG1}
\BIBentryALTinterwordspacing
{IPCC}, \emph{Climate Change 2021: The Physical Science Basis. Contribution of Working Group I to the Sixth Assessment Report of the Intergovernmental Panel on Climate Change}.\hskip 1em plus 0.5em minus 0.4em\relax Cambridge, United Kingdom and New York, NY, USA: Cambridge University Press, 2021, vol. In Press. [Online]. Available: \url{https://doi.org/10.1017/9781009157896}
\BIBentrySTDinterwordspacing

\bibitem{Neyshabur2017}
\BIBentryALTinterwordspacing
B.~Neyshabur, S.~Bhojanapalli, D.~Mcallester, and N.~Srebro, ``Exploring generalization in deep learning,'' in \emph{Advances in Neural Information Processing Systems}, I.~Guyon, U.~V. Luxburg, S.~Bengio, H.~Wallach, R.~Fergus, S.~Vishwanathan, and R.~Garnett, Eds., vol.~30.\hskip 1em plus 0.5em minus 0.4em\relax Curran Associates, Inc., 2017. [Online]. Available: \url{https://proceedings.neurips.cc/paper_files/paper/2017/file/10ce03a1ed01077e3e289f3e53c72813-Paper.pdf}
\BIBentrySTDinterwordspacing

\bibitem{Wang2022}
J.~Wang, C.~Lan, C.~Liu, Y.~Ouyang, T.~Qin, W.~Lu, Y.~Chen, W.~Zeng, and P.~Yu, ``Generalizing to unseen domains: {A} survey on domain generalization,'' \emph{IEEE Transactions on Knowledge and Data Engineering}, pp. 1--1, 2022.

\bibitem{shen2023engression}
X.~Shen and N.~Meinshausen, ``Engression: Extrapolation for nonlinear regression?'' 2023.

\bibitem{Roscher2020}
R.~Roscher, B.~Bohn, M.~Duarte, and J.~Garcke, ``Explainable machine learning for scientific insights and discoveries,'' \emph{IEEE Access}, vol.~PP, pp. 1--1, 02 2020.

\bibitem{Linardatos2021}
\BIBentryALTinterwordspacing
P.~Linardatos, V.~Papastefanopoulos, and S.~Kotsiantis, ``Explainable {AI}: {A} review of machine learning interpretability methods,'' \emph{Entropy}, vol.~23, no.~1, 2021. [Online]. Available: \url{https://www.mdpi.com/1099-4300/23/1/18}
\BIBentrySTDinterwordspacing

\bibitem{ras2022}
G.~Ras, N.~Xie, M.~Van~Gerven, and D.~Doran, ``Explainable deep learning: A field guide for the uninitiated,'' \emph{Journal of Artificial Intelligence Research}, vol.~73, pp. 329--396, 2022.

\bibitem{Mamalakis2022}
A.~Mamalakis, I.~Ebert-Uphoff, and E.~A. Barnes, \emph{Explainable Artificial Intelligence in Meteorology and Climate Science: Model Fine-Tuning, Calibrating Trust and Learning New Science}.\hskip 1em plus 0.5em minus 0.4em\relax Cham: Springer International Publishing, 2022, pp. 315--339.

\bibitem{hohl2024}
A.~H{\"o}hl, I.~Obadic, M.~{\'{A}}.~F. Torres, H.~Najjar, D.~Oliveira, Z.~Akata, A.~Dengel, and X.~X. Zhu, ``Opening the black-box: A systematic review on explainable ai in remote sensing,'' 2024.

\bibitem{Rudin2019}
\BIBentryALTinterwordspacing
C.~Rudin, ``Stop explaining black box machine learning models for high stakes decisions and use interpretable models instead,'' \emph{Nature Machine Intelligence}, vol.~1, no.~5, pp. 206--215, May 2019. [Online]. Available: \url{https://doi.org/10.1038/s42256-019-0048-x}
\BIBentrySTDinterwordspacing

\bibitem{Rudin2022}
\BIBentryALTinterwordspacing
C.~Rudin, C.~Chen, Z.~Chen, H.~Huang, L.~Semenova, and C.~Zhong, ``{Interpretable machine learning: Fundamental principles and 10 grand challenges},'' \emph{Statistics Surveys}, vol.~16, no. none, pp. 1 -- 85, 2022. [Online]. Available: \url{https://doi.org/10.1214/21-SS133}
\BIBentrySTDinterwordspacing

\bibitem{Sixt2020}
\BIBentryALTinterwordspacing
L.~Sixt, M.~Granz, and T.~Landgraf, ``When explanations lie: Why many modified {BP} attributions fail,'' in \emph{Proceedings of the 37th International Conference on Machine Learning}, ser. Proceedings of Machine Learning Research, H.~D. III and A.~Singh, Eds., vol. 119.\hskip 1em plus 0.5em minus 0.4em\relax PMLR, 13--18 Jul 2020, pp. 9046--9057. [Online]. Available: \url{https://proceedings.mlr.press/v119/sixt20a.html}
\BIBentrySTDinterwordspacing

\bibitem{freiesleben2023dear}
T.~Freiesleben and G.~K{\"o}nig, ``Dear xai community, we need to talk! fundamental misconceptions in current xai research,'' 2023.

\bibitem{karpatne2022knowledge}
A.~Karpatne, R.~Kannan, and V.~Kumar, \emph{{{Knowledge Guided Machine Learning}: Accelerating Discovery using Scientific Knowledge and Data}}, 1st~ed.\hskip 1em plus 0.5em minus 0.4em\relax Chapman and Hall/CRC, 2022.

\bibitem{CampsValls18sciasi}
G.~Camps-Valls, D.~Svendsen, L.~Martino, J.~Mu{\~{n}}oz-Mar{\'{i}}, V.~Laparra, M.~Campos-Taberner, and D.~Luengo, ``Physics-aware {G}aussian processes in remote sensing,'' \emph{Applied Soft Computing}, vol.~68, pp. 69--82, Jul 2018.

\bibitem{Tramontana2020}
\BIBentryALTinterwordspacing
G.~Tramontana, M.~Migliavacca, M.~Jung, M.~Reichstein, T.~F. Keenan, G.~Camps-Valls, J.~Ogee, J.~Verrelst, and D.~Papale, ``Partitioning net carbon dioxide fluxes into photosynthesis and respiration using neural networks,'' \emph{Global Change Biology}, vol.~26, no.~9, pp. 5235--5253, 2020. [Online]. Available: \url{https://onlinelibrary.wiley.com/doi/abs/10.1111/gcb.15203}
\BIBentrySTDinterwordspacing

\bibitem{khandelwal2020}
A.~Khandelwal, S.~Xu, X.~Li, X.~Jia, M.~Stienbach, C.~Duffy, J.~Nieber, and V.~Kumar, ``Physics guided machine learning methods for hydrology,'' 2020.

\bibitem{Cortes21fkl}
J.~Cort\'es-Andr\'es, G.~Camps-Valls, S.~Sippel, E.~Sz\'ekely, D.~Sejdinovic, E.~Diaz, A.~P\'erez-Suay, Z.~Li, M.~Mahecha, and M.~Reichstein, ``Physics-aware nonparametric regression models for {E}arth data analysis,'' \emph{Environmental Research Letters}, vol.~17, no.~5, 2022.

\bibitem{Liu2023}
L.~{LIU}, W.~{Zhou}, K.~{Guan}, B.~{Peng}, C.~{Jiang}, J.~{Tang}, S.~{Wang}, R.~{Grant}, S.~{Mezbahuddin}, X.~{Jia}, S.~{Xu}, V.~{Kumar}, and Z.~{Jin}, ``{Knowledge-based Artificial Intelligence for Agroecosystem Carbon Budget and Crop Yield Estimation},'' \emph{ESS Open Archive eprints}, vol. 105, p. essoar.10509206, Jul. 2023.

\bibitem{Zhu2022}
\BIBentryALTinterwordspacing
Q.~Zhu, F.~Li, W.~J. Riley, L.~Xu, L.~Zhao, K.~Yuan, H.~Wu, J.~Gong, and J.~Randerson, ``Building a machine learning surrogate model for wildfire activities within a global earth system model,'' \emph{Geoscientific Model Development}, vol.~15, no.~5, pp. 1899--1911, 2022. [Online]. Available: \url{https://gmd.copernicus.org/articles/15/1899/2022/}
\BIBentrySTDinterwordspacing

\bibitem{raissi2019}
M.~Raissi, P.~Perdikaris, and G.~Karniadakis, ``Physics-informed neural networks: {A} deep learning framework for solving forward and inverse problems involving nonlinear partial differential equations,'' \emph{Jour. Comp. Phys.}, vol. 378, pp. 686--707, 2019.

\bibitem{zhao2019physics}
W.~L. Zhao, P.~Gentine, M.~Reichstein, Y.~Zhang, S.~Zhou, Y.~Wen, C.~Lin, X.~Li, and G.~Y. Qiu, ``Physics-constrained machine learning of evapotranspiration,'' \emph{Geophysical Research Letters}, vol.~46, no.~24, pp. 14\,496--14\,507, 2019.

\bibitem{Reichstein2022}
M.~Reichstein, B.~Ahrens, B.~Kraft, G.~Camps-Valls, N.~Carvalhais, F.~Gans, P.~Gentine, and A.~J. Winkler, ``Combining system modeling and machine learning into hybrid ecosystem modeling,'' in \emph{Knowledge Guided Machine Learning}, 1st~ed.\hskip 1em plus 0.5em minus 0.4em\relax Chapman and Hall/CRC, 2022, p.~26.

\bibitem{Koppa2022}
\BIBentryALTinterwordspacing
A.~Koppa, D.~Rains, P.~Hulsman, R.~Poyatos, and D.~G. Miralles, ``A deep learning-based hybrid model of global terrestrial evaporation,'' \emph{Nature Communications}, vol.~13, no.~1, p. 1912, Apr 2022. [Online]. Available: \url{https://doi.org/10.1038/s41467-022-29543-7}
\BIBentrySTDinterwordspacing

\bibitem{Shen2023}
\BIBentryALTinterwordspacing
C.~Shen, A.~P. Appling, P.~Gentine, T.~Bandai, H.~Gupta, A.~Tartakovsky, M.~Baity-Jesi, F.~Fenicia, D.~Kifer, L.~Li, X.~Liu, W.~Ren, Y.~Zheng, C.~J. Harman, M.~Clark, M.~Farthing, D.~Feng, P.~Kumar, D.~Aboelyazeed, F.~Rahmani, Y.~Song, H.~E. Beck, T.~Bindas, D.~Dwivedi, K.~Fang, M.~H{\"o}ge, C.~Rackauckas, B.~Mohanty, T.~Roy, C.~Xu, and K.~Lawson, ``Differentiable modelling to unify machine learning and physical models for geosciences,'' \emph{Nature Reviews Earth {\&} Environment}, vol.~4, no.~8, pp. 552--567, Aug 2023. [Online]. Available: \url{https://doi.org/10.1038/s43017-023-00450-9}
\BIBentrySTDinterwordspacing

\bibitem{Oberpriller2021}
\BIBentryALTinterwordspacing
J.~Oberpriller, D.~R. Cameron, M.~C. Dietze, and F.~Hartig, ``Towards robust statistical inference for complex computer models,'' \emph{Ecology Letters}, vol.~24, no.~6, pp. 1251--1261, 2021. [Online]. Available: \url{https://onlinelibrary.wiley.com/doi/abs/10.1111/ele.13728}
\BIBentrySTDinterwordspacing

\bibitem{Abdar2021}
\BIBentryALTinterwordspacing
M.~Abdar, F.~Pourpanah, S.~Hussain, D.~Rezazadegan, L.~Liu, M.~Ghavamzadeh, P.~Fieguth, X.~Cao, A.~Khosravi, U.~R. Acharya, V.~Makarenkov, and S.~Nahavandi, ``A review of uncertainty quantification in deep learning: Techniques, applications and challenges,'' \emph{Information Fusion}, vol.~76, pp. 243--297, 2021. [Online]. Available: \url{https://www.sciencedirect.com/science/article/pii/S1566253521001081}
\BIBentrySTDinterwordspacing

\bibitem{Izmailov2021WhatAB}
\BIBentryALTinterwordspacing
P.~Izmailov, S.~Vikram, M.~D. Hoffman, and A.~G. Wilson, ``What are bayesian neural network posteriors really like?'' in \emph{International Conference on Machine Learning}, 2021. [Online]. Available: \url{https://api.semanticscholar.org/CorpusID:233443782}
\BIBentrySTDinterwordspacing

\bibitem{Kuhn2013}
M.~Kuhn and K.~Johnson, \emph{Applied Predictive Modeling}, 01 2013.

\bibitem{wang2022respecting}
S.~Wang, S.~Sankaran, and P.~Perdikaris, ``Respecting causality is all you need for training physics-informed neural networks,'' 2022.

\bibitem{iglesiassuarez2023causallyinformed}
F.~Iglesias-Suarez, P.~Gentine, B.~Solino-Fernandez, T.~Beucler, M.~Pritchard, J.~Runge, and V.~Eyring, ``Causally-informed deep learning to improve climate models and projections,'' 2023.

\bibitem{Runge19}
J.~Runge, S.~Bathiany, E.~Bollt, G.~Camps-Valls, D.~Coumou, E.~Deyle, C.~Clymour, M.~Kretschmer, M.~Mahecha, J.~Mu{\~{n}}oz-Mar{\'{i}}, E.~van Nes, J.~Peters, R.~Quax, M.~Reichstein, M.~Scheffer, B.~Sch{\"o}lkopf, P.~Spirtes, G.~Sugihara, J.~Sun, K.~Zhang, and J.~Zscheischler, ``Inferring causation from time series with perspectives in {E}arth system sciences,'' \emph{Nature Communications}, no. 2553, pp. 1--13, 2019.

\bibitem{YUAN2022109115}
\BIBentryALTinterwordspacing
K.~Yuan, Q.~Zhu, F.~Li, W.~J. Riley, M.~Torn, H.~Chu, G.~McNicol, M.~Chen, S.~Knox, K.~Delwiche, H.~Wu, D.~Baldocchi, H.~Ma, A.~R. Desai, J.~Chen, T.~Sachs, M.~Ueyama, O.~Sonnentag, M.~Helbig, E.-S. Tuittila, G.~Jurasinski, F.~Koebsch, D.~Campbell, H.~P. Schmid, A.~Lohila, M.~Goeckede, M.~B. Nilsson, T.~Friborg, J.~Jansen, D.~Zona, E.~Euskirchen, E.~J. Ward, G.~Bohrer, Z.~Jin, L.~Liu, H.~Iwata, J.~Goodrich, and R.~Jackson, ``Causality guided machine learning model on wetland ch4 emissions across global wetlands,'' \emph{Agricultural and Forest Meteorology}, vol. 324, p. 109115, 2022. [Online]. Available: \url{https://www.sciencedirect.com/science/article/pii/S0168192322003021}
\BIBentrySTDinterwordspacing

\bibitem{Chernozhukov2018}
\BIBentryALTinterwordspacing
V.~Chernozhukov, D.~Chetverikov, M.~Demirer, E.~Duflo, C.~Hansen, W.~Newey, and J.~Robins, ``Double/debiased machine learning for treatment and structural parameters,'' \emph{The Econometrics Journal}, vol.~21, no.~1, pp. C1--C68, 01 2018. [Online]. Available: \url{https://doi.org/10.1111/ectj.12097}
\BIBentrySTDinterwordspacing

\bibitem{Knaus_2020}
\BIBentryALTinterwordspacing
M.~C. Knaus, M.~Lechner, and A.~Strittmatter, ``Heterogeneous employment effects of job search programs,'' \emph{Journal of Human Resources}, vol.~57, no.~2, pp. 597--636, mar 2020. [Online]. Available: \url{https://doi.org/10.3368%2Fjhr.57.2.0718-9615r1}
\BIBentrySTDinterwordspacing

\bibitem{Davis2017}
\BIBentryALTinterwordspacing
J.~M. Davis and S.~B. Heller, ``Using causal forests to predict treatment heterogeneity: {A}n application to summer jobs,'' \emph{The American Economic Review}, vol. 107, no.~5, pp. 546--550, 2017. [Online]. Available: \url{http://www.jstor.org/stable/44250458}
\BIBentrySTDinterwordspacing

\bibitem{SUN2023}
\BIBentryALTinterwordspacing
Q.~Sun, T.~Zheng, X.~Zheng, M.~Cao, B.~Zhang, and S.~Jiang, ``Causal interpretation for groundwater exploitation strategy in a coastal aquifer,'' \emph{Science of The Total Environment}, vol. 867, p. 161443, 2023. [Online]. Available: \url{https://www.sciencedirect.com/science/article/pii/S004896972300058X}
\BIBentrySTDinterwordspacing

\bibitem{arrhenius1889reaktionsgeschwindigkeit}
S.~Arrhenius, ``{\"U}ber die reaktionsgeschwindigkeit bei der inversion von rohrzucker durch s{\"a}uren,'' \emph{Zeitschrift f{\"u}r physikalische Chemie}, vol.~4, no.~1, pp. 226--248, 1889.

\bibitem{van1899lectures}
J.~H. Van't~Hoff, R.~A. Lehfeldt \emph{et~al.}, ``Lectures on theoretical and physical chemistry,'' 1899.

\bibitem{lloyd1994temperature}
J.~Lloyd and J.~Taylor, ``On the temperature dependence of soil respiration,'' \emph{Functional ecology}, pp. 315--323, 1994.

\bibitem{PEI2022}
\BIBentryALTinterwordspacing
Y.~Pei, J.~Dong, Y.~Zhang, W.~Yuan, R.~Doughty, J.~Yang, D.~Zhou, L.~Zhang, and X.~Xiao, ``Evolution of light use efficiency models: Improvement, uncertainties, and implications,'' \emph{Agricultural and Forest Meteorology}, vol. 317, p. 108905, 2022. [Online]. Available: \url{https://www.sciencedirect.com/science/article/pii/S0168192322000983}
\BIBentrySTDinterwordspacing

\bibitem{kirschbaum2000will}
M.~U. Kirschbaum, ``Will changes in soil organic carbon act as a positive or negative feedback on global warming?'' \emph{Biogeochemistry}, vol.~48, pp. 21--51, 2000.

\bibitem{smith2013plant}
N.~G. Smith and J.~S. Dukes, ``Plant respiration and photosynthesis in global-scale models: incorporating acclimation to temperature and {CO2},'' \emph{Global change biology}, vol.~19, no.~1, pp. 45--63, 2013.

\bibitem{huntingford2017implications}
C.~Huntingford, O.~K. Atkin, A.~Martinez-De La~Torre, L.~M. Mercado, M.~A. Heskel, A.~B. Harper, K.~J. Bloomfield, O.~S. O'sullivan, P.~B. Reich, K.~R. Wythers \emph{et~al.}, ``Implications of improved representations of plant respiration in a changing climate,'' \emph{Nature Communications}, vol.~8, no.~1, p. 1602, 2017.

\bibitem{Vardi2023}
\BIBentryALTinterwordspacing
G.~Vardi, ``On the implicit bias in deep-learning algorithms,'' \emph{Commun. ACM}, vol.~66, no.~6, pp. 86--93, may 2023. [Online]. Available: \url{https://doi.org/10.1145/3571070}
\BIBentrySTDinterwordspacing

\bibitem{Zhan2022}
\BIBentryALTinterwordspacing
W.~Zhan, X.~Yang, Y.~Ryu, B.~Dechant, Y.~Huang, Y.~Goulas, M.~Kang, and P.~Gentine, ``Two for one: {P}artitioning {CO2} fluxes and understanding the relationship between solar-induced chlorophyll fluorescence and gross primary productivity using machine learning,'' \emph{Agricultural and Forest Meteorology}, vol. 321, p. 108980, 2022. [Online]. Available: \url{https://www.sciencedirect.com/science/article/pii/S0168192322001708}
\BIBentrySTDinterwordspacing

\bibitem{ElGhawi_2023}
\BIBentryALTinterwordspacing
R.~ElGhawi, B.~Kraft, C.~Reimers, M.~Reichstein, M.~K{\"o}rner, P.~Gentine, and A.~J. Winkler, ``Hybrid modeling of evapotranspiration: inferring stomatal and aerodynamic resistances using combined physics-based and machine learning,'' \emph{Environmental Research Letters}, vol.~18, no.~3, p. 034039, mar 2023. [Online]. Available: \url{https://dx.doi.org/10.1088/1748-9326/acbbe0}
\BIBentrySTDinterwordspacing

\bibitem{Yin_2021}
\BIBentryALTinterwordspacing
Y.~Yin, V.~L. Guen, J.~Dona, E.~de~B{\'e}zenac, I.~Ayed, N.~Thome, and P.~Gallinari, ``Augmenting physical models with deep networks for complex dynamics forecasting*,'' \emph{Journal of Statistical Mechanics: Theory and Experiment}, vol. 2021, no.~12, p. 124012, dec 2021. [Online]. Available: \url{https://dx.doi.org/10.1088/1742-5468/ac3ae5}
\BIBentrySTDinterwordspacing

\bibitem{Huenermund2021}
\BIBentryALTinterwordspacing
P.~H{\"u}nermund, B.~Louw, and I.~Caspi, ``Double machine learning and automated confounder selection: A cautionary tale,'' \emph{Journal of Causal Inference}, vol.~11, no.~1, p. 20220078, 2023. [Online]. Available: \url{https://doi.org/10.1515/jci-2022-0078}
\BIBentrySTDinterwordspacing

\bibitem{Athey2019}
\BIBentryALTinterwordspacing
S.~Athey, J.~Tibshirani, and S.~Wager, ``{Generalized random forests},'' \emph{The Annals of Statistics}, vol.~47, no.~2, pp. 1148 -- 1178, 2019. [Online]. Available: \url{https://doi.org/10.1214/18-AOS1709}
\BIBentrySTDinterwordspacing

\bibitem{Nie2020}
\BIBentryALTinterwordspacing
X.~Nie and S.~Wager, ``Quasi-oracle estimation of heterogeneous treatment effects,'' \emph{Biometrika}, vol. 108, no.~2, pp. 299--319, 09 2020. [Online]. Available: \url{https://doi.org/10.1093/biomet/asaa076}
\BIBentrySTDinterwordspacing

\bibitem{Foster2020}
D.~J. Foster and V.~Syrgkanis, ``Orthogonal statistical learning,'' 2020.

\bibitem{nekipelov2021regularized}
D.~Nekipelov, V.~Semenova, and V.~Syrgkanis, ``Regularized orthogonal machine learning for nonlinear semiparametric models,'' 2021.

\bibitem{Bonan_2015}
G.~Bonan, \emph{Ecological Climatology: Concepts and Applications}, 3rd~ed.\hskip 1em plus 0.5em minus 0.4em\relax Cambridge University Press, 2015.

\bibitem{Burba2013}
G.~Burba, \emph{Eddy Covariance Method for Scientific, Industrial, Agricultural and Regulatory Applications: {A} Field Book on Measuring Ecosystem Gas Exchange and Areal Emission Rates}, 06 2013.

\bibitem{Fluxnet}
D.~Baldocchi, E.~Falge, L.~Gu, R.~Olson, D.~Hollinger, S.~Running, P.~Anthoni, C.~Bernhofer, K.~Davis, R.~Evans, J.~Fuentes, A.~Goldstein, G.~Katul, B.~Law, X.~Lee, Y.~Malhi, T.~Meyers, W.~Munger, W.~Oechel, {Paw U,K.T.}, K.~Pilegaard, H.~Schmid, R.~Valentini, S.~Verma, T.~Vesala, K.~Wilson, and S.~Wofsy, ``\BIBforeignlanguage{English}{Fluxnet: A new tool to study the temporal and spatial variability of ecosystem-scale carbon dioxide, water vapor, and energy flux densities},'' \emph{\BIBforeignlanguage{English}{Bulletin of the American Meteorological Society}}, vol.~82, no.~11, pp. 2415--2434, 2001.

\bibitem{Falge2003}
\BIBentryALTinterwordspacing
E.~Falge, J.~Tenhunen, M.~Aubinet, C.~Bernhofer, R.~Clement, A.~Granier, A.~Kowalski, E.~Moors, K.~Pilegaard, {\"U}.~Rannik, and C.~Rebmann, \emph{A Model-Based Study of Carbon Fluxes at Ten European Forest Sites}.\hskip 1em plus 0.5em minus 0.4em\relax Berlin, Heidelberg: Springer Berlin Heidelberg, 2003, pp. 151--177. [Online]. Available: \url{https://doi.org/10.1007/978-3-662-05171-9_8}
\BIBentrySTDinterwordspacing

\bibitem{Pastorello2020}
G.~Pastorello, C.~Trotta, E.~Canfora, H.~Chu, D.~Christianson, Y.-W. Cheah, C.~Poindexter, J.~Chen, A.~Elbashandy, M.~Humphrey, P.~Isaac, D.~Polidori, M.~Reichstein, A.~Ribeca, C.~van Ingen, N.~Vuichard, L.~Zhang, B.~Amiro, C.~Ammann, M.~A. Arain, J.~Ard{\"o}, T.~Arkebauer, S.~K. Arndt, N.~Arriga, M.~Aubinet, M.~Aurela, D.~Baldocchi, A.~Barr, E.~Beamesderfer, L.~B. Marchesini, O.~Bergeron, J.~Beringer, C.~Bernhofer, D.~Berveiller, D.~Billesbach, T.~A. Black, P.~D. Blanken, G.~Bohrer, J.~Boike, P.~V. Bolstad, D.~Bonal, J.-M. Bonnefond, D.~R. Bowling, R.~Bracho, J.~Brodeur, C.~Br{\"u}mmer, N.~Buchmann, B.~Burban, S.~P. Burns, P.~Buysse, P.~Cale, M.~Cavagna, P.~Cellier, S.~Chen, I.~Chini, T.~R. Christensen, J.~Cleverly, A.~Collalti, C.~Consalvo, B.~D. Cook, D.~Cook, C.~Coursolle, E.~Cremonese, P.~S. Curtis, E.~D'Andrea, H.~da~Rocha, X.~Dai, K.~J. Davis, B.~D. Cinti, A.~d. Grandcourt, A.~D. Ligne, R.~C. De~Oliveira, N.~Delpierre, A.~R. Desai, C.~M. Di~Bella, P.~d. Tommasi, H.~Dolman, F.~Domingo, G.~Dong, S.~Dore,
  P.~Duce, E.~Dufr{\^e}ne, A.~Dunn, J.~Du{\v s}ek, D.~Eamus, U.~Eichelmann, H.~A.~M. ElKhidir, W.~Eugster, C.~M. Ewenz, B.~Ewers, D.~Famulari, S.~Fares, I.~Feigenwinter, A.~Feitz, R.~Fensholt, G.~Filippa, M.~Fischer, J.~Frank, M.~Galvagno, M.~Gharun, D.~Gianelle, B.~Gielen, B.~Gioli, A.~Gitelson, I.~Goded, M.~Goeckede, A.~H. Goldstein, C.~M. Gough, M.~L. Goulden, A.~Graf, A.~Griebel, C.~Gruening, T.~Gr{\"u}nwald, A.~Hammerle, S.~Han, X.~Han, B.~U. Hansen, C.~Hanson, J.~Hatakka, Y.~He, M.~Hehn, B.~Heinesch, N.~Hinko-Najera, L.~H{\"o}rtnagl, L.~Hutley, A.~Ibrom, H.~Ikawa, M.~Jackowicz-Korczynski, D.~Janou{\v s}, W.~Jans, R.~Jassal, S.~Jiang, T.~Kato, M.~Khomik, J.~Klatt, A.~Knohl, S.~Knox, H.~Kobayashi, G.~Koerber, O.~Kolle, Y.~Kosugi, A.~Kotani, A.~Kowalski, B.~Kruijt, J.~Kurbatova, W.~L. Kutsch, H.~Kwon, S.~Launiainen, T.~Laurila, B.~Law, R.~Leuning, Y.~Li, M.~Liddell, J.-M. Limousin, M.~Lion, A.~J. Liska, A.~Lohila, A.~L{\'o}pez-Ballesteros, E.~L{\'o}pez-Blanco, B.~Loubet, D.~Loustau, A.~Lucas-Moffat,
  J.~L{\"u}ers, S.~Ma, C.~Macfarlane, V.~Magliulo, R.~Maier, I.~Mammarella, G.~Manca, B.~Marcolla, H.~A. Margolis, S.~Marras, W.~Massman, M.~Mastepanov, R.~Matamala, J.~H. Matthes, F.~Mazzenga, H.~McCaughey, I.~McHugh, A.~M.~S. McMillan, L.~Merbold, W.~Meyer, T.~Meyers, S.~D. Miller, S.~Minerbi, U.~Moderow, R.~K. Monson, L.~Montagnani, C.~E. Moore, E.~Moors, V.~Moreaux, C.~Moureaux, J.~W. Munger, T.~Nakai, J.~Neirynck, Z.~Nesic, G.~Nicolini, A.~Noormets, M.~Northwood, M.~Nosetto, Y.~Nouvellon, K.~Novick, W.~Oechel, J.~E. Olesen, J.-M. Ourcival, S.~A. Papuga, F.-J. Parmentier, E.~Paul-Limoges, M.~Pavelka, M.~Peichl, E.~Pendall, R.~P. Phillips, K.~Pilegaard, N.~Pirk, G.~Posse, T.~Powell, H.~Prasse, S.~M. Prober, S.~Rambal, {\"U}.~Rannik, N.~Raz-Yaseef, C.~Rebmann, D.~Reed, V.~R.~d. Dios, N.~Restrepo-Coupe, B.~R. Reverter, M.~Roland, S.~Sabbatini, T.~Sachs, S.~R. Saleska, E.~P. S{\'a}nchez-Ca{\~n}ete, Z.~M. Sanchez-Mejia, H.~P. Schmid, M.~Schmidt, K.~Schneider, F.~Schrader, I.~Schroder, R.~L. Scott,
  P.~Sedl{\'a}k, P.~Serrano-Ort{\'\i}z, C.~Shao, P.~Shi, I.~Shironya, L.~Siebicke, L.~{\v S}igut, R.~Silberstein, C.~Sirca, D.~Spano, R.~Steinbrecher, R.~M. Stevens, C.~Sturtevant, A.~Suyker, T.~Tagesson, S.~Takanashi, Y.~Tang, N.~Tapper, J.~Thom, M.~Tomassucci, J.-P. Tuovinen, S.~Urbanski, R.~Valentini, M.~van~der Molen, E.~van Gorsel, K.~van Huissteden, A.~Varlagin, J.~Verfaillie, T.~Vesala, C.~Vincke, D.~Vitale, N.~Vygodskaya, J.~P. Walker, E.~Walter-Shea, H.~Wang, R.~Weber, S.~Westermann, C.~Wille, S.~Wofsy, G.~Wohlfahrt, S.~Wolf, W.~Woodgate, Y.~Li, R.~Zampedri, J.~Zhang, G.~Zhou, D.~Zona, D.~Agarwal, S.~Biraud, M.~Torn, and D.~Papale, ``The {FLUXNET2015} dataset and the {ONEFlux} processing pipeline for eddy covariance data,'' \emph{Scientific Data}, vol.~7, no.~1, p. 225, Jul. 2020.

\bibitem{Robinson1988}
\BIBentryALTinterwordspacing
P.~M. Robinson, ``Root-{N}-consistent semiparametric regression,'' \emph{Econometrica}, vol.~56, no.~4, pp. 931--954, 1988. [Online]. Available: \url{http://www.jstor.org/stable/1912705}
\BIBentrySTDinterwordspacing

\bibitem{kingma2017}
D.~P. Kingma and J.~Ba, ``Adam: {A} method for stochastic optimization,'' 2017.

\bibitem{Srivastava2014}
\BIBentryALTinterwordspacing
N.~Srivastava, G.~Hinton, A.~Krizhevsky, I.~Sutskever, and R.~Salakhutdinov, ``Dropout: {A} simple way to prevent neural networks from overfitting,'' \emph{Journal of Machine Learning Research}, vol.~15, no.~56, pp. 1929--1958, 2014. [Online]. Available: \url{http://jmlr.org/papers/v15/srivastava14a.html}
\BIBentrySTDinterwordspacing

\bibitem{Krogh1991}
A.~Krogh and J.~A. Hertz, ``A simple weight decay can improve generalization,'' in \emph{Proceedings of the 4th International Conference on Neural Information Processing Systems}, ser. NIPS'91.\hskip 1em plus 0.5em minus 0.4em\relax San Francisco, CA, USA: Morgan Kaufmann Publishers Inc., 1991, pp. 950--957.

\bibitem{Luo2006}
Y.~Luo and X.~Zhou, ``Soil respiration and the environment,'' \emph{Soil Respiration and the Environment}, 01 2006.

\bibitem{Chapin2013-ko}
F.~S. Chapin, P.~A. Matson, and H.~A. Mooney, \emph{\BIBforeignlanguage{en}{Principles of terrestrial ecosystem ecology}}, 2002nd~ed.\hskip 1em plus 0.5em minus 0.4em\relax New York, NY: Springer, May 2013.

\bibitem{Reichstein2005}
\BIBentryALTinterwordspacing
M.~Reichstein, E.~Falge, D.~Baldocchi, D.~Papale, M.~Aubinet, P.~Berbigier, C.~Bernhofer, N.~Buchmann, T.~Gilmanov, A.~Granier, T.~Gr{\"u}nwald, K.~Havr{\'{a}}nkov{\'{a}}, H.~Ilvesniemi, D.~Janous, A.~Knohl, T.~Laurila, A.~Lohila, D.~Loustau, G.~Matteucci, T.~Meyers, F.~Miglietta, J.-M. Ourcival, J.~Pumpanen, S.~Rambal, E.~Rotenberg, M.~Sanz, J.~Tenhunen, G.~Seufert, F.~Vaccari, T.~Vesala, D.~Yakir, and R.~Valentini, ``On the separation of net ecosystem exchange into assimilation and ecosystem respiration: review and improved algorithm,'' \emph{Global Change Biology}, vol.~11, no.~9, pp. 1424--1439, 2005. [Online]. Available: \url{https://onlinelibrary.wiley.com/doi/abs/10.1111/j.1365-2486.2005.001002.x}
\BIBentrySTDinterwordspacing

\bibitem{Moffat2007}
\BIBentryALTinterwordspacing
A.~M. Moffat, D.~Papale, M.~Reichstein, D.~Y. Hollinger, A.~D. Richardson, A.~G. Barr, C.~Beckstein, B.~H. Braswell, G.~Churkina, A.~R. Desai, E.~Falge, J.~H. Gove, M.~Heimann, D.~Hui, A.~J. Jarvis, J.~Kattge, A.~Noormets, and V.~J. Stauch, ``Comprehensive comparison of gap-filling techniques for eddy covariance net carbon fluxes,'' \emph{Agricultural and Forest Meteorology}, vol. 147, no.~3, pp. 209--232, 2007. [Online]. Available: \url{https://www.sciencedirect.com/science/article/pii/S016819230700216X}
\BIBentrySTDinterwordspacing

\bibitem{Desai2008}
A.~R. Desai, A.~D. Richardson, A.~M. Moffat, J.~Kattge, D.~Hollinger, A.~G. Barr, E.~Falge, A.~Noormets, D.~Papale, M.~Reichstein, and V.~J. Stauch, ``Cross-site evaluation of eddy covariance {GPP} and {RE} decomposition techniques,'' \emph{Agricultural and Forest Meteorology}, vol. 148, pp. 821--838, 2008.

\bibitem{Lasslop2010}
\BIBentryALTinterwordspacing
G.~Lasslop, M.~Reichstein, D.~Papale, A.~D. Richardson, A.~Arneth, A.~Barr, P.~Stoy, and G.~Wohlfahrt, ``Separation of net ecosystem exchange into assimilation and respiration using a light response curve approach: critical issues and global evaluation,'' \emph{Global Change Biology}, vol.~16, no.~1, pp. 187--208, 2010. [Online]. Available: \url{https://onlinelibrary.wiley.com/doi/abs/10.1111/j.1365-2486.2009.02041.x}
\BIBentrySTDinterwordspacing

\bibitem{Keenan2019}
\BIBentryALTinterwordspacing
T.~F. Keenan, M.~Migliavacca, D.~Papale, D.~Baldocchi, M.~Reichstein, M.~Torn, and T.~Wutzler, ``Widespread inhibition of daytime ecosystem respiration,'' \emph{Nature Ecology {\&} Evolution}, vol.~3, no.~3, pp. 407--415, Mar 2019. [Online]. Available: \url{https://doi.org/10.1038/s41559-019-0809-2}
\BIBentrySTDinterwordspacing

\bibitem{Trifunov2021}
V.~T. Trifunov, M.~Shadaydeh, J.~Runge, M.~Reichstein, and J.~Denzler, ``A data-driven approach to partitioning net ecosystem exchange using a deep state space model,'' \emph{IEEE Access}, vol.~9, pp. 107\,873--107\,883, 2021.

\bibitem{Friedman2021}
\BIBentryALTinterwordspacing
J.~H. Friedman, ``{Greedy function approximation: A gradient boosting machine.}'' \emph{The Annals of Statistics}, vol.~29, no.~5, pp. 1189 -- 1232, 2001. [Online]. Available: \url{https://doi.org/10.1214/aos/1013203451}
\BIBentrySTDinterwordspacing

\bibitem{gal16}
\BIBentryALTinterwordspacing
Y.~Gal and Z.~Ghahramani, ``Dropout as a {B}ayesian approximation: {R}epresenting model uncertainty in deep learning,'' in \emph{Proceedings of The 33rd International Conference on Machine Learning}, ser. Proceedings of Machine Learning Research, M.~F. Balcan and K.~Q. Weinberger, Eds., vol.~48.\hskip 1em plus 0.5em minus 0.4em\relax New York, New York, USA: PMLR, 20--22 Jun 2016, pp. 1050--1059. [Online]. Available: \url{https://proceedings.mlr.press/v48/gal16.html}
\BIBentrySTDinterwordspacing

\bibitem{Mahecha2010}
\BIBentryALTinterwordspacing
M.~D. Mahecha, M.~Reichstein, N.~Carvalhais, G.~Lasslop, H.~Lange, S.~I. Seneviratne, R.~Vargas, C.~Ammann, M.~A. Arain, A.~Cescatti, I.~A. Janssens, M.~Migliavacca, L.~Montagnani, and A.~D. Richardson, ``Global convergence in the temperature sensitivity of respiration at ecosystem level,'' \emph{Science}, vol. 329, no. 5993, pp. 838--840, 2010. [Online]. Available: \url{https://www.science.org/doi/abs/10.1126/science.1189587}
\BIBentrySTDinterwordspacing

\bibitem{Scott2015}
\BIBentryALTinterwordspacing
R.~L. Scott, J.~A. Biederman, E.~P. Hamerlynck, and G.~A. Barron-Gafford, ``The carbon balance pivot point of southwestern u.s. semiarid ecosystems: Insights from the 21st century drought,'' \emph{Journal of Geophysical Research: Biogeosciences}, vol. 120, no.~12, pp. 2612--2624, 2015. [Online]. Available: \url{https://agupubs.onlinelibrary.wiley.com/doi/abs/10.1002/2015JG003181}
\BIBentrySTDinterwordspacing

\bibitem{Jung2020}
\BIBentryALTinterwordspacing
M.~Jung, C.~Schwalm, M.~Migliavacca, S.~Walther, G.~Camps-Valls, S.~Koirala, P.~Anthoni, S.~Besnard, P.~Bodesheim, N.~Carvalhais, F.~Chevallier, F.~Gans, D.~S. Goll, V.~Haverd, P.~K\"ohler, K.~Ichii, A.~K. Jain, J.~Liu, D.~Lombardozzi, J.~E. M.~S. Nabel, J.~A. Nelson, M.~O'Sullivan, M.~Pallandt, D.~Papale, W.~Peters, J.~Pongratz, C.~R\"odenbeck, S.~Sitch, G.~Tramontana, A.~Walker, U.~Weber, and M.~Reichstein, ``Scaling carbon fluxes from eddy covariance sites to globe: synthesis and evaluation of the fluxcom approach,'' \emph{Biogeosciences}, vol.~17, no.~5, pp. 1343--1365, 2020. [Online]. Available: \url{https://bg.copernicus.org/articles/17/1343/2020/}
\BIBentrySTDinterwordspacing

\end{thebibliography}

\newpage
\appendix
\section{Method}
\subsection{Derivation of \ac{DML} estimator for $g$}\label{sec: g estimator}
One way of obtaining an estimator for $g$ instead of fitting it directly is by reusing all estimators of \ac{DML}. It is easy to see that
\begin{align*}
    g(X,W) &= \E[g(X,W)|X,W]\\
            &= \E[Y-\theta(X)f(T)-\epsilon|X,W]\\
              &= \E[Y|X,W] - \E[\theta(X)f(T)|X,W] -\underbrace{\E[\epsilon|X,W]}_{=0}\\
              &= \E[Y|X,W] - \theta(X)\E[f(T)|X,W]\\
              &\approx \E[Y|X,W] - \hat{\theta}(X)\E[f(T)|X,W],
\end{align*}
where $\E[Y|X,W]$ represents the estimator of $Y$ on $X$ and $W$ and $\E[f(T)|X,W]$ the estimator of $f(T)$ on $X$ and $W$. From here, one can use an ensemble of the first-stage estimators over all folds to obtain the estimator of $\E[Y|X,W]$ and the estimator of $\E[f(T)|X,W]$. The estimator $\hat{\theta}(X)$ is a single estimator obtained as the result of \ac{DML}.

\section{Data}
\subsection{Synthetic data}
\subsubsection{$\Q$ model} \label{sec: syn Q10 model}
We use measured air temperature $T_A$ and potential incoming radiation $SW_{POT}$ for the synthetic data. Further, we compute
\begin{align}
    &\text{for }\Q \in \{1.5,1.25,1.75\},\\
    &\RECOsyn = \Rbsyn \cdot \Q^{0.1\cdot(T_A-15)}\cdot (1 + \epsilon),\\
    &\Rbsyn = 0.75 \cdot (\tilde{R}^{syn}_b - \min(\tilde{R}^{syn}_b) + 0.1\cdot \pi),\label{eq:Rb_zero}\\
    &\tilde{R}^{syn}_b = 0.01 \cdot \SWPOTsm - 0.005 \cdot \SWPOTsmdiff,
\end{align}
where $\Rb^{syn}$ describes the base respiration, which we compute with a smooth daily radiation cycle. The smooth incoming potential radiation $\SWPOTsm$ and its smoothed difference quotient $\SWPOTsmdiff$ are computed by averaging moving windows of 10 days over the incoming potential radiation $SW_{POT}$. We apply the computations in \eqref{eq:Rb_zero} to ensure that $\Rbsyn$ is always positive. We sample $\epsilon$ from a centered truncated normal distribution with $0.2$ standard deviation in the interval $[-0.95,0.95]$ to obtain heteroscedastic noise over the observations.

\subsubsection{$\LUE$ model} \label{sec: syn LUE model}
The code for generating the data is taken from the work of \cite{Reichstein2022}, where the authors approach the partitioning of fluxes with neural networks on a synthetic dataset.
$\RECOsyn$ is computed similarly as in the study on $\Q$.
While, for generating $GPP$, we use the light-use efficiency model with $\LUE$ being a function of $\VPD$ and temperature $T_A$:
\begin{align}
    \GPP^{syn} &= \LUE^{syn} \cdot \SW_{in},\\
    \LUE^{syn} &= 0.5 \cdot \exp \left( -0.1\cdot (T_A-20))^2\right) \cdot \min(1,\exp(-0.1\cdot (\VPD - 10))).
\end{align}
Finally, we compute $NEE$ following 
\eqref{eq:nee}
with additional multiplicative heteroscedastic noise:
\begin{align}
    NEE^{syn} &= (-GPP^{syn}+\RECOsyn)\cdot (1+\sigma \varepsilon),
\end{align}
where noise $\varepsilon\sim \mathcal{N}(0,1)$ is sampled from a standard Gaussian distribution and $\sigma$ varies in $\{0, 0.05, 0.1, 0.2, 0.4, 0.7, 1.0, 2.0\}$.

\subsection{FLUXNET sites}\label{sec: sites}
The 36 FLUXNET sites used for the flux partitioning experiments are shown in ~\cref{tbl:flxnet}. The table further provides information on plant type, latitude, and longitude.

\begin{table}
\centering
\caption{FLUXNET sites used for flux partitioning experiments with \ac{DML}.}
\label{tbl:flxnet}
\begin{tabular}{llllll}
\bottomrule
          ID & Site code &  IGBP & Lat & Lon & Years available\\
\midrule
        1 & AU-Cpr & SAV & -34,00 & 140,59  & 2010--2014\\
        2 & AU-DaP & GRA & -14,06 & 131,32 & 2007--2013\\
        3 & AU-Dry & SAV & -15,26 & 132,37 & 2008--2014\\
        4 & AU-How & WSA & -12,49 & 131,15 & 2001--2014\\
        5 & AU-Stp & GRA & -17,15 & 133,35  & 2008--2014\\
        6 & BE-Lon & CRO & 50,55 & 4,75 & 2004--2014\\
        7 & BE-Vie & MF & 50,31 & 6,00 & 1996--2014\\
        8 & CA-Qfo & ENF & 49,69 & -74,34  & 2003--2010\\
        9 & DE-Geb & CRO & 51,10 & 10,91  & 2001--2014\\
        10 & DE-Gri & GRA & 50,95 & 13,51  & 2004--2014\\
        11 & DE-Kli & CRO & 50,89 & 13,52  & 2004--2014\\
        12 & DE-Obe & ENF & 50,79 & 13,72  & 2008--2014\\
        13 & DE-Tha & ENF & 50,96 & 13,57  & 1996--2014\\
        14 & DK-Sor & DBF & 55,49 & 11,64  & 1996--2014\\
        15 & FI-Hyy & ENF & 61,85 & 24,29 & 1996--2014\\
        16 & FR-LBr & ENF & 44,72 & -0,77 & 1996--2008\\
        17 & GF-Guy & EBF & 5,28 & -52,92 & 2004--2014\\
        18 & IT-BCi & CRO & 40,52 & 14,96 & 2004--2014\\
        19 & IT-Cp2 & EBF & 41,70 & 12,36 & 2012--2014\\
        20 & IT-Cpz & EBF & 41,71 & 12,38 & 1997--2009\\
        21 & IT-MBo & GRA & 46,01 & 11,05 & 2003--2013\\
        22 & IT-Noe & CSH & 40,61 & 8,15 & 2004--2014\\
        23 & IT-Ro1 & DBF & 42,41 & 11,93 & 2000--2008\\
        24 & IT-SRo & ENF & 43,73 & 10,28 & 1999--2012\\
        25 & NL-Loo & ENF & 52,17 & 5,74 & 1996--2014\\
        26 & RU-Fyo & ENF & 56,46 & 32,92 & 1998--2014\\
        27 & US-ARM & CRO & 36,61 & -97,49 & 2003--2012\\
        28 & US-GLE & ENF & 41,37 & -106,24 & 2004--2014\\
        29 & US-MMS & DBF & 39,32 & -86,41 & 1999--2014\\
        30 & US-NR1 & ENF & 40,03 & -105,55 & 1999--2014\\
        31 & US-SRG & GRA & 31,79 & -110.83 & 2008--2014\\
        32 & US-SRM & WSA & 31,82 & -110,87 & 2004--2014\\
        33 & US-UMB & DBF & 45,56 & -84,71 & 2000--2014\\
        34 & US-Whs & OSH & 31,74 & -110,05 & 2007--2014\\
        35 & US-Wkg & GRA & 31,74 & -109,94 & 2004--2014\\
        36 & ZA-Kru & SAV & -25,02 & 31,50 & 2000--2013\\
\bottomrule
\end{tabular}
\end{table}

\subsection{Details on the neural networks}\label{sec: model details}
The \acp{DNN} used for the \ac{GDHM} had two hidden layers with 16 units each. A $\tanh$ nonlinearity was applied at the end of each hidden layer. A final softplus function was applied to the output of the last layer to obtain non-negative results for the base respiration. This function is a smooth approximation of the $ReLU$ function. For the case of regularization, dropout was applied to the outputs of the hidden layers at a rate of $0.2$. To probe other instances of regularization, we also used weight decay with hyperparameter $0.1$ instead of dropout. The initial $\Q$ is sampled from a Gaussian with $\sigma=0.1$ and $\mu=1.5$ (or $1.25$, $1.75$ for the respective experiments).
For the \ac{DMLHM} approach, we used the same network architecture without final softplus for the first-stage estimators. For the estimation of $\Rb$ after obtaining $\Q$, we used the same network again, but this time we included the softplus nonlinearity.
We used stochastic gradient descent with the Adam optimizer~\cite{kingma2017} for the training. We apply exponential learning rate decay as a scheduler with a decay rate of $0.95$ over $500$ steps. We trained the first stage estimators of the \ac{DML} over $2000$ iterations each. For the \ac{GDHM} and the final $g$ estimator in the causal \ac{DMLHM}, we trained over $10000$ iterations. To avoid overfitting, $20\%$ of the data is always kept as validation data for model selection.

\section{Additional results}
\subsection{Regularization with weight decay}
\begin{figure}
    \centering
     \includegraphics[width=0.5\textwidth]{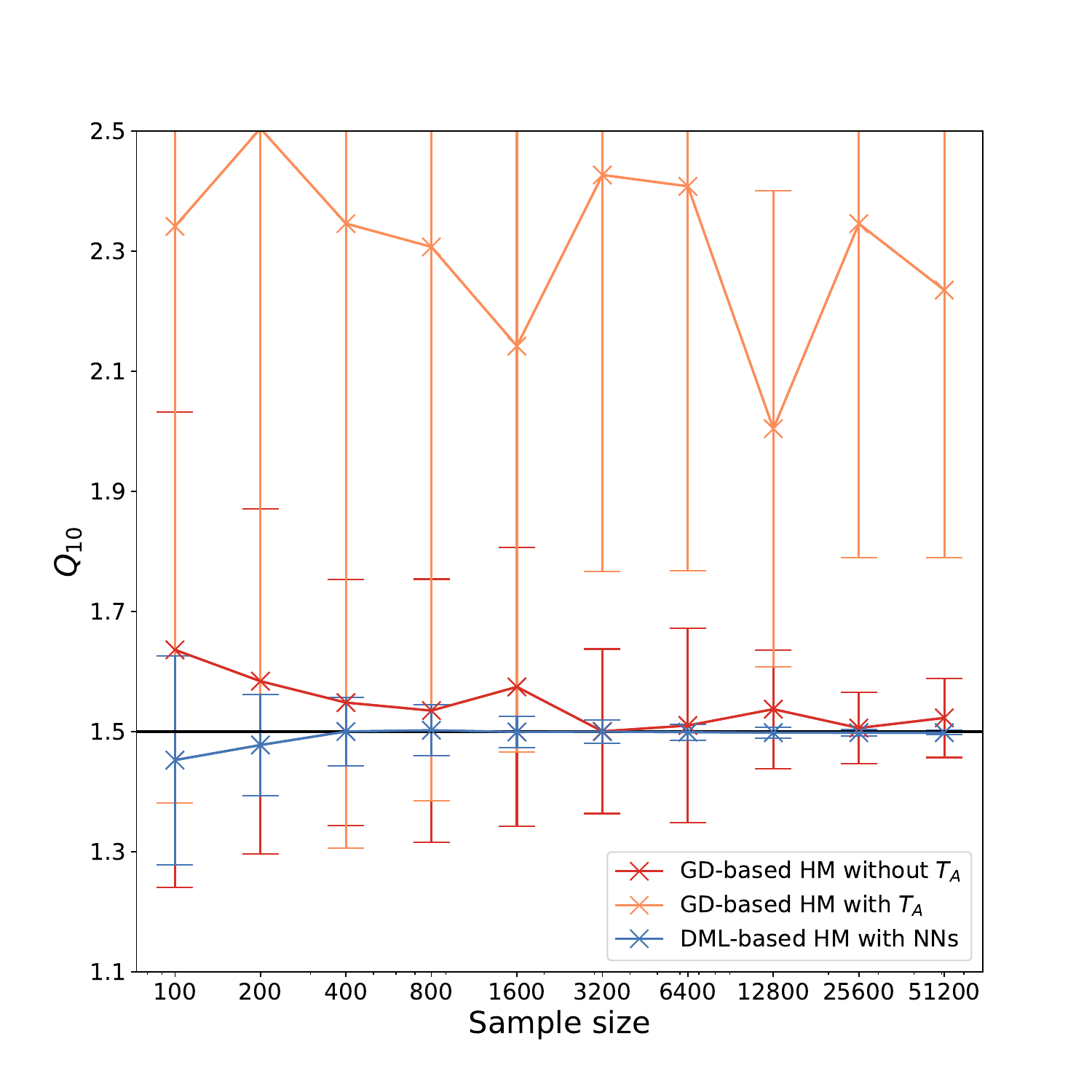}
  \caption{
      Additional simulation study for $\Q$ estimation with the \ac{GDHM} and the \ac{DMLHM} similar to \cref{fig: Q10}. with weight decay. Here, weight decay with a rate of $0.1$ has been applied as regularization.}
  \label{fig: Q10 with weight decay}
\end{figure}

We reran the same setup with weight decay to show that the findings also apply to other regularization techniques beyond dropout. We find qualitatively similar results, where the \ac{DMLHM} converges robustly to the right $\Q$ values where the \ac{GDHM} converges much slower and remains biased (see \cref{fig: Q10 with weight decay}).
\label{sec: weight decay}

\subsection{Additional $\Q$ values.}
\label{sec: additional Q10 values}
\begin{figure}
\label{fig: additional Q10 values}
  \begin{subfigure}{0.5\textwidth}
    \centering
     \includegraphics[width=1.0\textwidth]{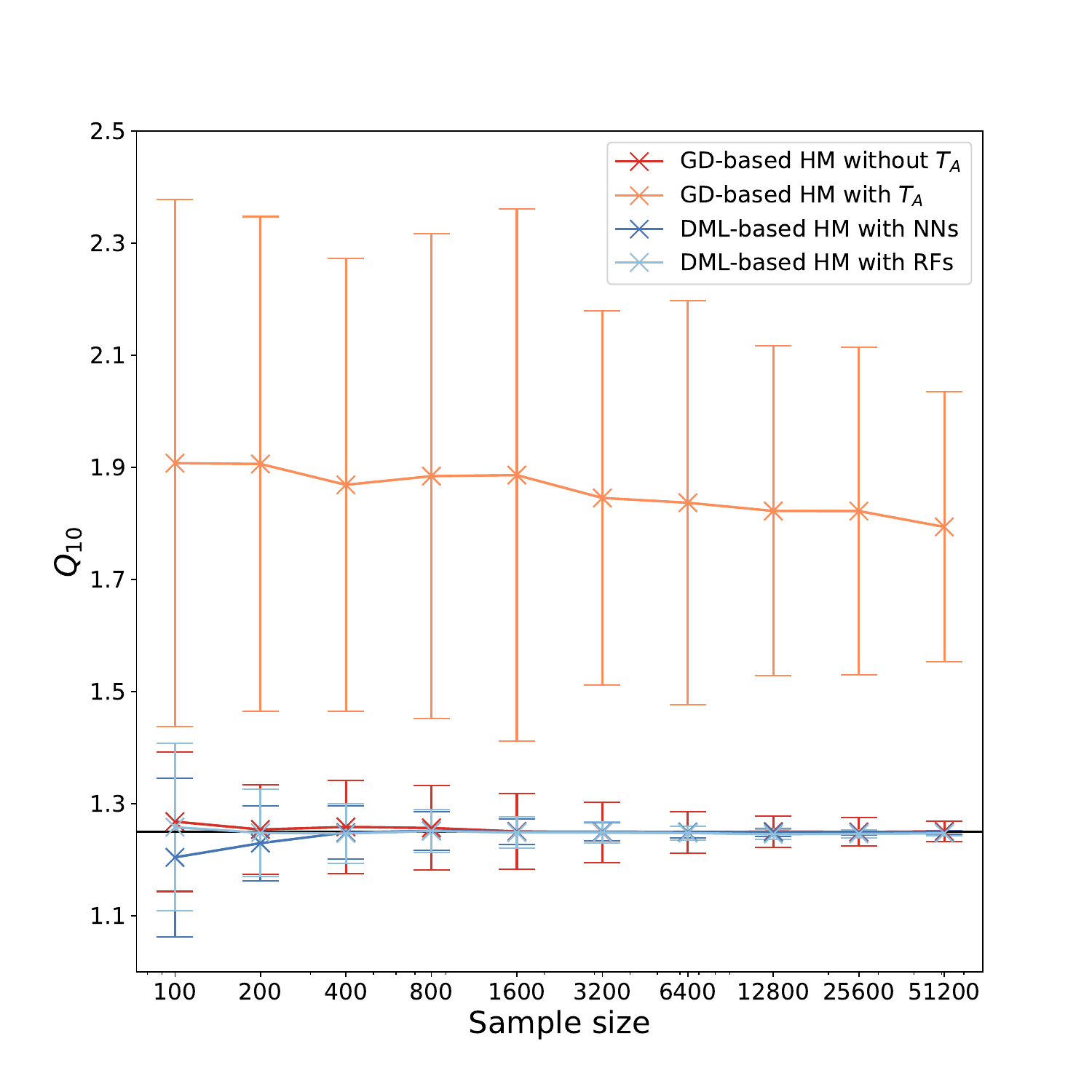}
    \caption{$\Q$ of 1.25 without dropout.}
    \label{fig: 1.25 without dropout}
  \end{subfigure}
    \begin{subfigure}{0.5\textwidth}
    \centering
     \includegraphics[width=1.0\textwidth]{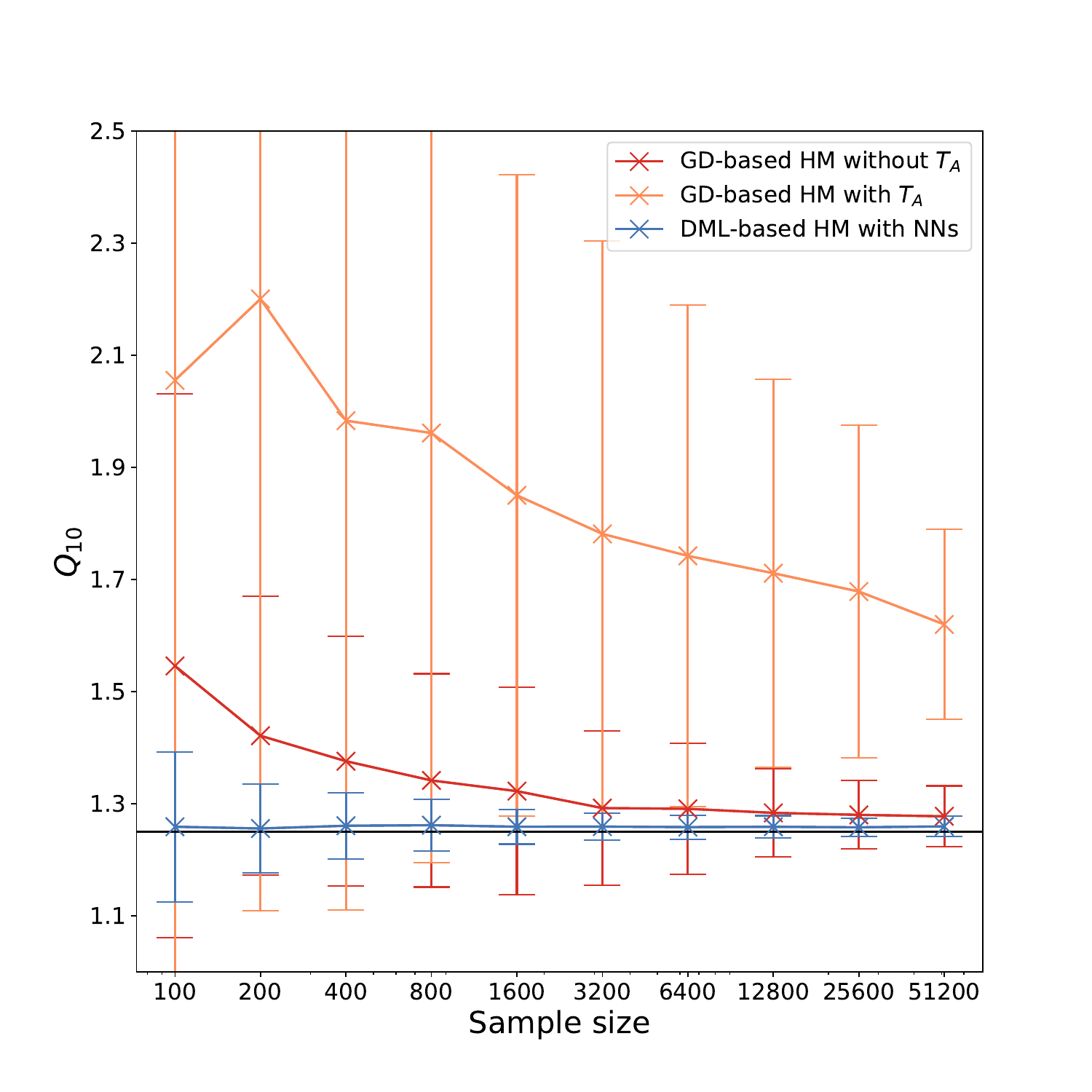}
    \caption{$\Q$ of 1.25 with dropout.}
    \label{fig: 1.25 with dropout}
  \end{subfigure}
    \begin{subfigure}{0.5\textwidth}
    \centering
     \includegraphics[width=1.0\textwidth]{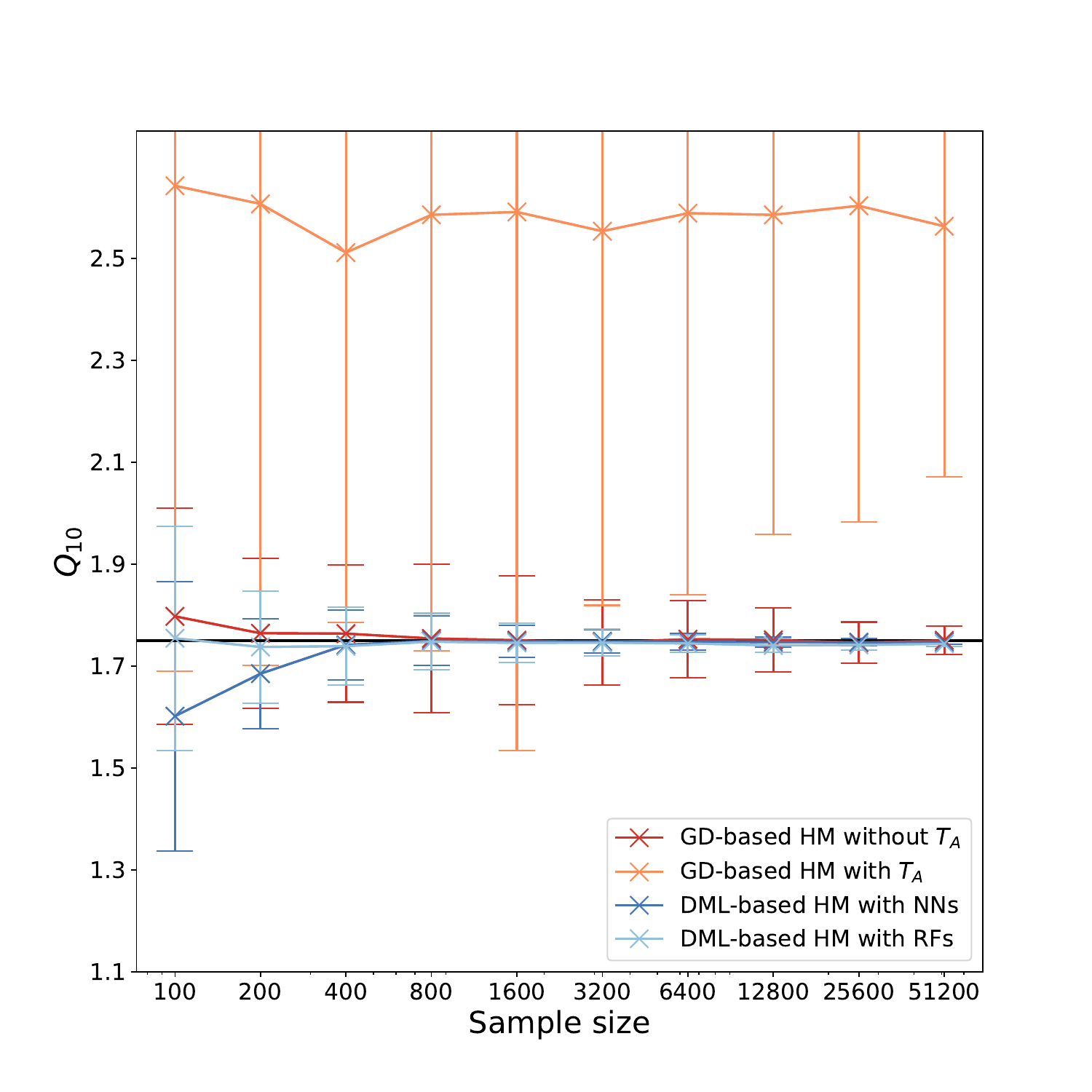}
    \caption{$\Q$ of 1.75 without dropout.}
    \label{fig: 1.75 without dropout}
  \end{subfigure}
    \begin{subfigure}{0.5\textwidth}
    \centering
     \includegraphics[width=1.0\textwidth]{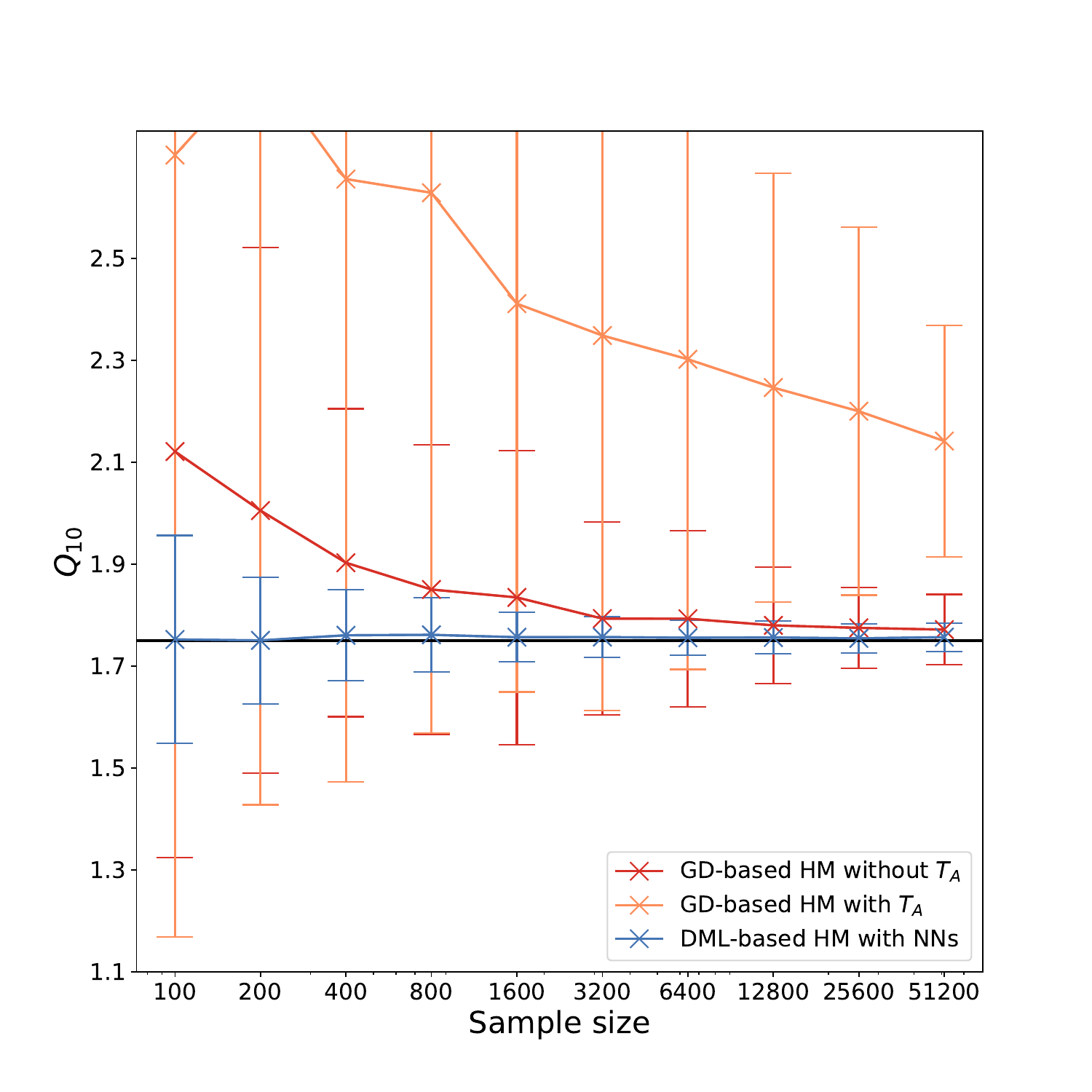}
    \caption{$\Q$ of 1.75 with dropout.}
    \label{fig: 1.75 with dropout}
  \end{subfigure}
  \caption{
      Additional simulation study for $\Q$ estimation with the \ac{GDHM} and the \ac{DMLHM} similar to \cref{fig: Q10} with different values for $\Q$ In a) and b) $\Q$ was set to $1.25$ and in c) and d) to $1.75$. The findings are qualitatively similar to the case of $1.5$. The magnitude of the errors grows with the magnitude of $\Q$.}
\end{figure}

We ran the experiments with and without dropout with $1.25$ and $1.75$ as two additional $\Q$ values. We find that these setups affirm the observations for $\Q=1.5$. The errors in estimating the $\Q$ values grow and shrink proportionally to the magnitude of $\Q$. This is to be expected as we deploy multiplicative noise, and thus, with higher $\Q$, the magnitude of respiration and, hence, the absolute noise level grows (see \cref{fig: additional Q10 values}).

\subsection{Retrievel of linear model}\label{sec: linear flux partitioning}
We generated synthetic data following~\cite{Reichstein2022}, a partially linear $LUE$ model with varying coefficients. We used time series of measured meteorological forcings as inputs and added heteroscedastic noise over different noise levels (see \cref{sec: syn LUE model} for details). 

To test the robustness of the approach to noise, 
we perform experiments with an increasing level of heteroscedastic noise.
The $R^2$ and \ac{RMSE} of the retrieved fluxes are reported in \cref{tbl:results_syn_R2} and \cref{tbl:results_syn_RMSE}. 
We note that the \ac{DML} approach gives theoretical guarantees for estimating $\GPP$ and not necessarily for $\RECO$~\cite{Athey2019, Foster2020}. 
Our proposed method retrieves reasonable estimates of $\GPP$ with a medium $R^2$ of $0.997$ in the no-noise scenario. Even a heteroscedastic noise level of 
$0.4$ does not yield any substantial drop in performance. 
Beyond that, the method is still robust as it retrieves the correct $\GPP$ at a noise level of $1.00$ with a median value of $0.922$. 
In flux partitioning, retrieving $\RECO$ can be more challenging as it has a smaller magnitude than $\GPP$, implying a smaller signal-to-noise ratio. Moreover, even though there is no guarantee on the used plugin-in estimator for $\RECO$, which we obtain by recycling the estimators of the \ac{DML} approach, we still find it to yield useful results. The retrieved fluxes have a median $R^2$ over all site-years of $0.94$. As expected, the effect of the noise on the retrieval of $\RECO$ is stronger, but up to a $\sigma$ of $0.4$, the results are not strongly affected. When we combine both models, we obtain a model of $\NEE$. Even with strong noise, this estimator retrieves reasonable estimates of the $\NEE$ signal.

\begin{table}
\centering
\caption{Coefficient of determination $R^2$ for generated data on all 36 flux sites with different heteroscedastic noise levels between the \ac{GPP}, \ac{RECO} and \ac{NEE} obtained with the \ac{DML} approach and the respective ground truth. For \ac{NEE}, the noise-free value is stated. The reported statistics are the median and in brackets, the 0.25 and 0.75 quantiles over all site-years.}
\label{tbl:results_syn_R2}
\begin{tabular}{RLLL}
\toprule
\sigma &               \GPP &              \RECO &         \NEE_{clean}  \\
\midrule
       0.00 &  0.997 (0.994/0.998) &    0.940 (0.923/0.960) &  0.978 (0.973/0.983) \\
       0.05 &  0.997 (0.994/0.998) &   0.940 (0.923/0.959) &  0.978 (0.973/0.983) \\
       0.10 &  0.997 (0.993/0.998) &  0.939 (0.922/0.958) &  0.978 (0.973/0.982) \\
       0.20 &  0.996 (0.991/0.998) &  0.936 (0.917/0.956) &  0.977 (0.972/0.982) \\
       0.40 &  0.993 (0.985/0.996) &  0.931 (0.911/0.947) &  0.975 (0.969/0.979) \\
       0.70 &  0.986 (0.961/0.991) &  0.914 (0.888/0.929) &   0.970 (0.963/0.975) \\
       1.00 &   0.977 (0.930/0.985) &   0.887 (0.846/0.910) &   0.964 (0.955/0.970) \\
       2.00 &  0.922 (0.707/0.952) &  0.751 (0.617/0.813) &   0.937 (0.910/0.948) \\
\bottomrule
\end{tabular}
\end{table}
\begin{table}
\centering
\caption{The \ac{RMSE} (in $\frac{\mu \mol \ce{CO2}}{m^{2}s}$) for generated data on all 36 flux sites with different heteroscedastic noise levels between the \ac{GPP}, \ac{RECO} and \ac{NEE} obtained with the \ac{DML} approach and the respective ground truth. For \ac{NEE}, the noise-free and noisy values are stated. The reported statistics are the median and, in brackets, the 0.25 and 0.75 quantiles over all site-years.}
\label{tbl:results_syn_RMSE}
\begin{tabular}{RLLLL}
\toprule
\sigma &               \GPP &              \RECO &      \NEE_{clean} &   \NEE_{noisy} \\
\midrule
       0.00 &  0.320 (0.227/0.454) &  0.861 (0.770/1.104) &  0.872 (0.768/1.079) &  \phantom{0}0.872 (\phantom{0}0.768/\phantom{0}1.079) \\
       0.05 &  0.330 (0.234/0.467) &  0.864 (0.771/1.109) &  0.873 (0.770/1.083) &  \phantom{0}1.029 (\phantom{0}0.827/\phantom{0}1.311) \\
       0.10 &  0.359 (0.243/0.491) &  0.878 (0.778/1.136) &  0.880 (0.770/1.097) &  \phantom{0}1.197 (\phantom{0}0.949/\phantom{0}1.615) \\
       0.20 &  0.401 (0.284/0.600) &  0.921 (0.794/1.184) &  0.898 (0.781/1.128) &  \phantom{0}1.701 (\phantom{0}1.346/\phantom{0}2.573) \\
       0.40 &  0.515 (0.386/0.772) &  0.973 (0.825/1.335) &  0.941 (0.808/1.219) &  \phantom{0}2.977 (\phantom{0}2.349/\phantom{0}4.850) \\
       0.70 &  0.758 (0.543/1.152) &  1.139 (0.895/1.577) &  1.025 (0.862/1.358) &  \phantom{0}5.101 (\phantom{0}3.965/\phantom{0}8.434) \\
       1.00 &  1.005 (0.715/1.589) &  1.285 (0.971/1.872) &  1.147 (0.927/1.467) &  \phantom{0}7.162 (\phantom{0}5.583/11.949) \\
       2.00 &  1.804 (1.268/2.972) &  1.880 (1.361/3.058) &  1.500 (1.196/2.186) &  14.316 (11.104/23.889) \\
\bottomrule
\end{tabular}
\end{table}

\section{Reproducibility}
The data used to carry out experiments is available at \url{https://fluxnet.org/data/fluxnet2015-dataset/}. All code is being made available at \url{https://github.com/KaiHCohrs/hybrid-q10-model-chm} and \url{https://github.com/KaiHCohrs/dml-4-fluxes-chm}.

\end{document}